%% file: automl2026_conference.tex
\documentclass{article} 
\usepackage{amsmath}
\usepackage{amssymb}
\usepackage{amsthm}
\usepackage[final]{automl}

\newcommand{\review}[1]{{\leavevmode\color{black}#1}}

\usepackage{url}

\usepackage{microtype} 
\usepackage{booktabs}  
\usepackage{url}  
\usepackage{csquotes}
\usepackage[dvipsnames]{xcolor}
\usepackage[ruled, vlined, linesnumbered, commentsnumbered, longend]{algorithm2e} 
\SetArgSty{textnormal}
\usepackage{algorithmic}
\usepackage[ruled, vlined, linesnumbered, commentsnumbered, longend]{algorithm2e} 
\SetArgSty{textnormal}
\usepackage{algorithmic}
\usepackage{graphicx}
\usepackage{tabularx}
\usepackage[export]{adjustbox}
\usepackage{witharrows}
\usepackage{tikz}
\usepackage{forest}
\usepackage{bm}
\newcommand{\offspring}[0]{o\hspace{-0.05em}f\hspace{-0.12em}f\hspace{-0.06em}spring}
\usepackage{todonotes}

\usepackage{dsfont}
\usepackage{pifont}
\usepackage{float}
\usepackage{enumitem}

\usepackage{multicol}

\usepackage[subtle]{savetrees}

\usepackage[numbers,square]{natbib}
\newcommand{\cmark}{\textcolor{green}{\ding{51}}} 
\newcommand{\xmark}{\textcolor{red}{\ding{55}}}   

\expandafter\def\expandafter\normalsize\expandafter{%
    \normalsize%
    \setlength\abovedisplayskip{6pt}%
    \setlength\belowdisplayskip{6pt}%
    \setlength\abovedisplayshortskip{-6pt}%
    \setlength\belowdisplayshortskip{2pt}%
}

\input{math_commands.tex}

\title{Evolutionary Architecture Search \\ Through Grammar-Based Sequence Alignment}

\usepackage{authblk}
\author[1,2,*]{Adri Gómez}
\author[3,*]{Felix Möller}
\author[4]{Steven McDonagh}
\author[1,2]{Monica Abella}
\author[1,2]{Manuel Desco}
\author[4]{Elliot J. Crowley}
\author[5]{Aaron Klein}
\author[6]{Linus Ericsson}

\affil[ ]{%
  \begin{minipage}[t]{0.60\textwidth}
    \raggedright
    \textsuperscript{1}Carlos III University of Madrid \newline
    \textsuperscript{2}Instituto de Investigación Sanitaria Gregorio Marañón \newline
    \textsuperscript{3}Helmholtz-Zentrum Berlin für Materialien und Energie
  \end{minipage}%
  \hfill
  \begin{minipage}[t]{0.291\textwidth}
    \raggedright
    \textsuperscript{4}University of Edinburgh \newline
    \textsuperscript{5}ELLIS Institute Tübingen \newline
    \textsuperscript{6}University of Glasgow \newline
    \textsuperscript{*}{\emph{Equal contribution}}
  \end{minipage}%
}

\begin{document}

\maketitle

\begin{abstract}
Neural architecture search (NAS) in expressive search spaces is a computationally hard problem, but it also holds the potential to automatically discover completely novel and performant architectures.
Effective recombination first requires establishing correspondence between related substructures in the parent architectures so that architectural components can be reused in new candidates.
We formulate this correspondence as a grammar-constrained sequence-alignment problem and introduce two adaptations of the Smith-Waterman dynamic-programming framework.
The resulting grammar-valid edit paths define an architecture distance and generate hybrid offspring from two parent models, supporting crossover-based search, architectural loss-landscape analysis, and population-diversity tracking.
We highlight how our method vastly improves computational complexity over previous work and enables us to efficiently compute shortest paths between architectures.
When instantiating the crossover in evolutionary searches, we obtain competitive results relative to the evaluated baselines while enabling substantially richer analysis and control of the evolution.
Future work can build upon this new tool, discovering novel components that can be used more broadly across neural architecture design, and broadening its applications beyond NAS. Our code can be found at: \url{https://github.com/flxai/rcswx-paper}.
\end{abstract}

\section{Introduction}
Neural architecture search (NAS)~\citep{elsken-iclr19} has traditionally operated in constrained search spaces with limited operations and fixed topologies.
Benchmarks such as NAS-Bench-101~\citep{ying-icml19} and NAS-Bench-201~\citep{dong-iclr20} make optimisation tractable by restricting architectures to small cell-based templates, but this also limits the possibility of discovering fundamentally new designs.
Recent grammar-based spaces address this limitation by allowing variable depth, branching, routing, and operation types; examples include Hierarchical NAS~\citep{schrodi2023constructionhierarchicalneuralarchitecture} and \textit{einspace}~\citep{einspace}.
This expressiveness creates a new algorithmic challenge: search operators and distance metrics designed for small DAG-based spaces do not directly scale to flexible grammar-derived architectures.
Evolutionary NAS is especially affected by this mismatch.
Mutation remains easy to define, but crossover and architecture distances must account for large structured encodings, syntactic validity, and permutation-equivalent representations.
Shortest edit path crossover (SEPX) addresses recombination by finding graph edits that transform one parent into another while respecting such equivalences~\citep{qiu2023shortest}.
However, exact graph edit distance is NP-hard~\citep{bougleux2017graph}, and SEPX becomes impractical for larger architectures with many nodes and connections, as shown in Figure~\ref{fig:cswx-sepx-times}.

We propose a scalable alternative based on grammar-aware local sequence alignment.
Using the context-free grammar representation of \textit{einspace}, we serialise each architecture into a token sequence and adapt the Smith-Waterman algorithm~\citep{smith1981identification} to compute grammar-valid edit paths between parent architectures.
These edit paths induce a distance metric, guide recombination, and support population-diversity tracking and architectural loss-landscape analysis.
A recursive variant additionally handles branch-order permutations, yielding a distance over functional architectures rather than only over their serialised encodings.
The alignment has a second role beyond recombination: its edit cost provides a tractable architecture distance, which enables quantitative analysis of local landscape structure and population diversity in expressive grammar-based spaces where exact graph-edit measures are prohibitively expensive.
Our contributions are:
\begin{itemize}
    \item We introduce constrained and recursive constrained Smith-Waterman crossover operators for grammar-based NAS, producing syntactically valid edit paths
    and architecture distances.
    \item We show that these operators reduce computation by orders of magnitude relative to graph-edit-based SEPX while enabling crossover, diversity tracking, and loss-landscape analysis in one of the most expressive NAS spaces studied to date.
    \item We use the resulting RCSWX distance to analyse local landscape smoothness and population diversity in two search trajectories, each comprising 1,000 architectures, demonstrating an analytical use of the method beyond offspring generation.
\end{itemize}

\section{Background and Related Work}

NAS methods are commonly described by a search space, a search strategy, and a performance estimator~\citep{elsken-iclr19}.
This work focuses on search strategies for expressive grammar-based spaces, where architectures are generated by production rules rather than selected from a fixed cell template.
In \textit{einspace}~\citep{einspace}, rules such as \texttt{M -> B M M A} define branching modules in which two independently sampled submodules are combined by a branching and an aggregation operator.
Such rules make it possible to represent CNN-like, Transformer-like, routing, and branching structures in a shared space, but they also introduce non-sequential structure and branch-order permutation symmetries.

Evolutionary NAS maintains a population of candidate architectures and applies variation operators such as mutation and crossover~\citep{EvoNASreview}.
Subtree crossover (STX) naturally applies to derivation trees by exchanging non-terminal subtrees~\citep{STX_book}, but it does not account for functionally equivalent branch permutations and relies on mutation for much of the fine-grained exploration.
Ordered tree edit distance methods~\citep{Zhang1989SimpleFA} have the same limitation.
Graph-edit-based methods such as SEPX~\citep{qiu2023shortest} address permutation invariance by recombining parents along a shortest edit path, but their reliance on exact graph edit distance restricts them to small graphs in practice.

Sequence-alignment methods offer a scalable alternative by comparing linearised architecture encodings with dynamic programming.
Needleman-Wunsch and Smith-Waterman alignment~\citep{NW,smith1981identification} have been used to compute distances and control population diversity in sequential NAS spaces~\citep{pava2024sequence}, but these methods do not directly handle branching, skip-like structure, or permutation-equivalent encodings.
Our method combines sequence alignment with grammar constraints and recursive handling of branch permutations, retaining the crossover and distance-metric benefits of edit-path methods while scaling to richer architecture topologies.

\section{Proposed Method}

Given a grammar-derived architecture, we serialise its derivation tree into a token sequence and compute grammar-valid edit paths between parent architectures.
We propose two variants.
Constrained Smith-Waterman crossover (CSWX) applies local sequence alignment with validity constraints, producing syntactically valid edit paths, distances, and hybrid offspring.
Recursive CSWX (RCSWX) additionally handles branch-order permutations by recursively aligning permutable subsections, yielding a distance over functional architectures rather than only over their serialised encodings.
The resulting operators provide the crossover and distance-metric functionality of shortest edit path methods while avoiding the cost of exact graph edit distance by exploiting the grammar structure.
For further explanation of subtree crossover (STX) and shortest edit path crossover (SEPX), see Appendix~\ref{appendix:alt_methods}.

\subsection{Constrained Smith-Waterman Crossover}

\begin{figure}[H]
\hspace{-0.05\linewidth}
\begin{minipage}[t]{0.55\linewidth}

\centering
\captionsetup{width=0.83\linewidth}
\includegraphics[
                 width=0.9\linewidth, valign=t]{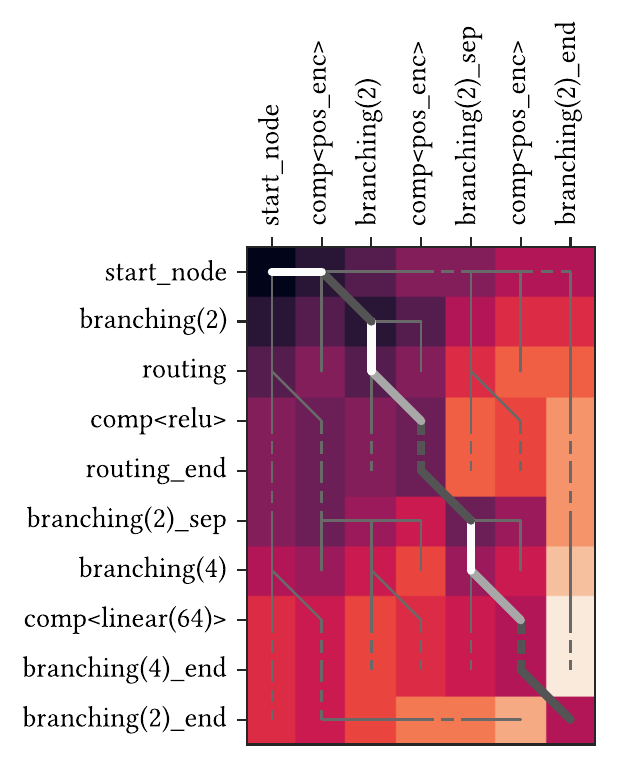}
\caption{Example $dists$ and $paths$ matrices overlaid.
Lighter coloured cells denote a higher distance from the first parent model, shown on top.
Moving towards the right entails deleting a node from the first model, moving downwards represents the addition of a node from the second model, and moving diagonally corresponds to a node substitution.
Dashed lines signal the closure of branching and routing nodes.
The optimal mutation path is traced back in thicker lines, whose brightness reflects the weight of each operation, being brightest intensities a cost of 1, and darker ones reaching a cost of 0.}
\label{fig:matrix}

\end{minipage}%
\begin{minipage}[t]{0.5\linewidth}
\vspace{-9.75pt}

Then, we compute the minimum-cost path that transforms the first parent into the second via additions, deletions, and substitutions.
For each operation considered, we check its validity through the \textbf{ValidPath} function.
This ensures that we only substitute nodes of the same type (e.g., a branching node cannot be replaced by a terminal node).
Moreover, if we try to add, delete, or substitute a separator node, it checks that we performed the same operation to the corresponding Routing or Branching node, and that all Routing and Branching nodes we may have added, deleted, or substituted within that path are properly closed by their corresponding separator, avoiding incongruous models.
We sequentially fill all positions within the $dist$ and $paths$ matrices, starting from the top left, selecting the path that presents the smallest cost to reach each one.
Once the bottom right corner of the matrix is reached, the best path is traced back and transformed into a series of operations through the \textbf{TraceBack} function.
Following said path from Figure~\ref{fig:matrix} would yield the following offspring.


\vspace{7pt}
\includegraphics[width=1\linewidth]{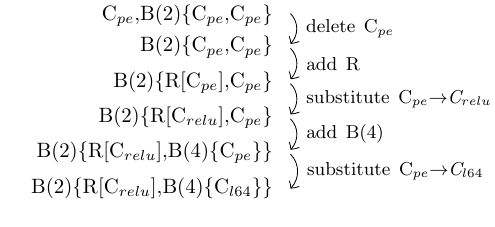}
\vspace{-19.5pt}

If we performed all these operations, we would simply obtain the second model;
as we are interested in creating a hybrid offspring, we sample a subset of operations through the \textbf{SelectOperations} function.

\end{minipage}
\end{figure}

We introduce CSWX in Algorithm~\ref{alg:CSWC}.
Using \textbf{Serialise}, we convert each parent model into a simplified sequence.
Nodes implied by structure, e.g. \emph{Sequential}, \emph{Grouping}, \emph{Aggregation}, are omitted.
To delimit branches and routing nodes, we insert $separator$ tokens.
To satisfy Smith-Waterman, we prefix each sequence with a $start$ token and place the two sequences on perpendicular axes (see Figure~\ref{fig:matrix}).

\vspace{-2pt}

\vspace{-8pt}
Lastly, these operations are applied within the \textbf{GenerateOffspring} function, which produces the desired offspring as the same tree-based format used for the parent models.
For visualisations of the architectures that can be constructed along the shortest path in Figure~\ref{fig:matrix}, see Appendix~\ref{appendix:intermediate_offspring}. Further details and pseudocode for all described functions is provided in Appendix \ref{appendix:algs}.


\vspace{0pt}
\subsection{Recursive Constrained Smith-Waterman Crossover}
Collapsing each architecture to a serialised sequence that is fed to CSWX introduces a branch-order permutation problem.
As an example, a skip connection around a linear layer $(B(2)\{C_{l64}, C_{id}\})$ is functionally identical regardless of the order of the linear layer and the identity operation in the encoding.
Computing the complete alignment matrix for every combination of branch swaps in either parent architecture becomes
intractable as the number of permutations increases for longer models.
We therefore propose a recursive version of CSWX: RCSWX, which reuses all possible pre-computed information in the alignment matrix, and only recalculates the cells that would be affected by the swapping of the branches.
RCSWX scales much better on bigger architectures while identifying the same set of operations for generating hybrid offspring as the brute force approach.



RCSWX separates the alignment matrix into submatrices delimited by branching nodes. The number of nested branching nodes for a certain submatrix is referred to as its depth $d$.
For each submatrix, we enumerate the $2^d$ possible branch-order permutations as auxiliary submatrices and compute them as in CSWX.
At each $separator$ token that closes a branching node, RCSWX collapses the corresponding submatrices into one by retaining, for every cell, the minimum-cost path across all permutations.
Note that the collapse does not imply selecting a singular submatrix, but rather combining them by selecting the best path to reach each individual cell in the alignment matrix.
Collapsing at the $separator$ tokens enforces the constraints and preserves global optimality, yielding permutation invariance while keeping computation feasible.

\begin{algorithm}
{\small
\caption{Constrained Smith-Waterman Crossover}\label{alg:CSWC}

\KwIn{\hspace{10pt}$model1$, the derivation tree for the first architecture to cross over\\
\hspace{40pt}$model2$, the derivation tree for the second architecture to cross over}
\hspace{40pt}$skewness$, the skewness of the operation sampling probability distribution\\
\KwOut{\hspace{2pt}$\offspring$, the derivation tree for the resulting hybrid architecture}
\vspace{10pt}
$model1_{seq}  \gets$ \textbf{Serialise}$(model1)$\\
$model2_{seq} \gets$ \textbf{Serialise}$(model2)$\\
$dist \gets$ empty array of dimensions length of $model1_{seq}$ $\times$ length of $model2_{seq}$\\
$paths \gets$ empty array of dimensions length of $model1_{seq}$ $\times$ length of $model2_{seq}$\\
\For{$i \leftarrow 1$ \KwTo length of $model1_{seq}$}{
    \For{$j \leftarrow 1$ \KwTo length of $model2_{seq}$}{
        $sub,$ $add,$ $del \gets \infty$\\
        \If{\textbf{ValidPath}$(paths,$ $model1_{seq},$ $model2_{seq},$ $i,$ $j,$ "sub"$)$}
        {
        $sub \gets dist[i-1,j-1] +$ \textbf{SubstitutionCost}$(model1_{seq}[i],$ $model2_{seq}[j])$
        }
        \If{\textbf{ValidPath}$(paths,$ $model1_{seq},$ $model2_{seq},$ $i,$ $j,$ "add"$)$}
        {
        $add \gets dist[i-1,j] + 1 - (model1_{seq}[i]$ is a separator node)
        }
        \If{\textbf{ValidPath}$(paths,$ $model1_{seq},$ $model2_{seq},$ $i,$ $j,$ "del"$)$}
        {
        $del \gets dist[i,j-1] + 1 - (model2_{seq}[j]$ is a separator node)
        }
        $dist[i,j] \gets \min(sub,$ $add,$ $del)$\\
        $paths[i,j] \gets
        \left(
        \textnormal{``sub''},
        \textnormal{``add''},
        \textnormal{``del''}
        \right)
        \left[\arg\min(sub,add,del)\right]$
        }
    }
    
$ops_{valid} \gets$ \textbf{TraceBack}$(model1_{seq},$ $model2_{seq},$ $dist,$ $paths)$\\
$ops_{selected} \gets$ \textbf{SelectOperations}$(ops_{valid},$ $skewness)$\\
$\offspring \gets$ \textbf{GenerateOffspring}$(model1_{seq},$ $model2_{seq},$ $ops_{selected})$\\
}
\end{algorithm}

\subsection{Properties of CSWX and RCSWX}
\subsubsection{Asymptotic Scaling}\label{scaling}
The SEPX algorithm~\citep{qiu2023shortest} requires 
computing 
the Graph Edit Distance (GED) between the 
parent architectures, which finds the node correspondence that minimises the total edit cost.
For two graphs of size $n_1$ and $n_2$, such a combinatorial search yields $O\left(n_1^{n_2}\right)$ scaling, which makes it
intractable for crossover of longer architectures.
In contrast, by treating the architectures as serialised trees instead of graphs we can use dynamic programming through the Smith--Waterman algorithm, which gives us CSWX that scales with
\begin{equation}
T_{CSWX} = O\left(n_1 n_2\right).
\end{equation}
Making CSWX permutation invariant by enumerating all permutations of $b$ branching nodes introduces an exponential factor $2^b$, yielding a total scaling of $O\!\left(n_1 n_2 2^b\right)$ for this brute-force approach.
Considering that each branching block is composed of 5 tokens (opening, first branch, separator, second branch, closing), then in the worst case every branch can itself be a branching block.
This yields a maximum of $b = (n_1+n_2)/4$ branch nodes, giving worst-case scaling
\begin{equation}
T_{BF-CSWX} = O\!\left(n_1 n_2\,2^{\frac{n_1+n_2}{4}}\right).
\end{equation}



RCSWX reduces this exponential factor by reusing partial results within each alignment cell.
Instead of a global cost of $2^p$ (with $p$ the maximum nesting depth), the cost locally at a cell $(i,j)$ is only $2^{d_{ij}}$, where $d_{ij}$ is the number of simultaneously open branches.
The total runtime is therefore $\sum_{i=1}^{n_1}\sum_{j=1}^{n_2} 2^{d_{ij}}$.
The worst case scenario, where both parent models are composed of maximally nested branching blocks along a single branch---which is similar to what may happen in U-Net style networks~\citep{unet}---yields a scaling of 
\begin{equation}
T_{RCSWX} = O\left(m2^{2m}\right)
\end{equation}
for $m=\frac{n_1}{4}-1$ assuming $n_1=n_2$.
RCSWX is faster than SEPX and brute-force CSWX in all cases.

\subsubsection{Memory usage}
CSWX stores an $n_1 \times n_2$ alignment matrix together with the traceback information needed to recover a grammar-valid edit path.
RCSWX reduces repeated computation by additionally retaining auxiliary alignment states for the branch-order permutations that are active at a given point in the recursion and collapsing these states when the corresponding branch blocks close.
It therefore trades additional memory for reduced recomputation:
peak memory depends not only on the serialized sequence lengths, but also on the number and nesting depth of simultaneously active branches and on the alignment states retained for traceback.
This can become a practical limitation for very large or deeply nested architectures.
Earlier collapsing of equivalent states, path deduplication, and banded alignment are possible ways to reduce this memory footprint;
Appendix~\ref{appendix:big_models} discusses related implementation considerations.

\subsubsection{CSWX and RCSWX as Distance Metrics}\label{metrics_definition}
The edit path computed by (R)CSWX naturally yields an edit distance between two architectures.
We consider this distance on two related domains.
Let $\mathcal{A}$ denote the set of neural architectures, where two architectures are considered identical if they are functionally equivalent (e.g., differing only by functionally inconsequential permutations of operations).
Let $\mathcal{B}$ denote the set of syntactic representations of these architectures in the \textit{einspace} grammar (sequences or trees, where branch order is explicit).
We now show that the edit distances defined by CSWX and RCSWX satisfy the axioms of a metric on their respective domains.

\paragraph{Non-negativity} Each edit operation has a cost $c_{i,j}\in[0,1]$, so summing over them yields nonnegative edit distances.

\paragraph{Identity of Indiscernibles} If two input architectures are identical, the shortest edit path follows the diagonal of the alignment matrix with all mutation costs $c_{i,j}=0$, giving $d_{\text{CSWX}}(x,x)=0$ for any $x \in \mathcal{B}$.
For non-identical syntactic representations, CSWX may assign a positive cost even when the architectures are functionally equivalent (e.g., differing only by branch permutations).
RCSWX resolves this by mapping syntactic forms in $\mathcal{B}$ to their functional counterparts in $\mathcal{A}$, ensuring that
\begin{equation}
d_{\text{RCSWX}}(x,y) = 0 \;\;\iff\;\; f_{\mathcal{B}\rightarrow\mathcal{A}}(x) = f_{\mathcal{B}\rightarrow\mathcal{A}}(y), \quad x,y \in \mathcal{B}.
\end{equation}

\paragraph{Symmetry} Swapping the two input architectures simply transposes the alignment matrix.
All node additions will become node deletions and vice-versa, and the final distance will be identical.

\paragraph{Triangle Inequality}  
The edit distance between two architectures is defined as the cost of the optimal alignment path between them.
Consider three architectures $x,y,z$.
The path from $x$ to $z$ can be decomposed by first aligning $x$ to $y$, and then $y$ to $z$.
Concatenating these two valid edit paths yields a feasible (though not necessarily optimal) path from $x$ to $z$ with cost $d(x,y)+d(y,z)$.
Since Smith--Waterman always finds the minimal-cost valid alignment under the grammar constraints, the optimal cost from $x$ to $z$ cannot exceed this concatenated cost.
Therefore,
\begin{equation}
d(x,z) \;\leq\; d(x,y) + d(y,z).
\end{equation}

\medskip
By satisfying all axioms, $d_{\text{CSWX}}: \mathcal{B} \times \mathcal{B} \to \mathbb{R}_{\geq 0}$ is a metric in the \emph{syntactic} space, while $d_{\text{RCSWX}}: \mathcal{A} \times \mathcal{A} \to \mathbb{R}_{\geq 0}$ is a metric in the \emph{semantic} space of architectures.

\section{Experiments} \label{sec:exps}
This section lays out the experiments performed to check the capabilities of the (R)CSWX algorithm, both in terms of exploration capabilities and computational expense.
We use a subset of datasets from the Unseen NAS benchmark~\citep{unseenNAS}.
Population sizes, running times and other similar hyperparameters were based off of previous experience on the datasets and explored search spaces~\citep{unseenNAS,einspace}.
Experiments at scale
ran on JUWELS~\citep{kesselheim2021juwelsboostersupercomputer};
code for these original experiments is available under the MIT license at \url{https://github.com/flxai/rcswx-paper}.

We evaluate three distinct questions.
Section~\ref{exp1} tests whether CSWX and RCSWX remain competitive when used as full-scale evolutionary variation operators, using STX and mutation-only search as baselines that can complete the full evaluation budget.
SEPX is the closest conceptual baseline for permutation-aware edit-path crossover, but its graph-edit computation becomes intractable at the architecture sizes encountered in these searches.
Section~\ref{scalability} therefore compares RCSWX directly with SEPX on tractable architecture pairs in terms of edit-path agreement and runtime.
Section~\ref{smoothness} evaluates the separate use of RCSWX as an architecture-distance measure.

\subsection{Search Performance Analysis}\label{exp1}
We compare the results yielded by (R)CSWX with those obtained using Subtree Crossover (STX) and 
with no crossover across \review{five} different random seeds. 
SEPX becomes intractable at the architecture sizes typical of our search, making a direct search performance comparison infeasible.
We set both the crossover and mutation probabilities to 1.0; these are offspring-level event probabilities.
The no-crossover condition disables crossover and otherwise uses the same mutation procedure.
After crossover, a single mutation is applied by sampling one terminal or non-terminal node uniformly from the offspring and replacing the selected node with a newly sampled grammar-valid subtree.
When a terminal node is selected, its current operation is excluded from the available choices before resampling.
Invalid mutation proposals are resampled subject to the implementation's configured retry limit of 100.
All (R)CSWX hyperparameters, including the substitution weights and operation-sampling skewness, are left at their default values;
see Appendix~\ref{appendix:algs}.

We start with a randomly sampled initial population of 100 architectures and run the search for 900 further iterations, resulting in 1,000 architecture evaluations.
Each update follows a steady-state regularised-evolution procedure:
one offspring is generated and evaluated, appended to the population, and the chronologically oldest individual is removed.
Each parent is selected using a tournament of size 10.
Tournament candidates are sampled uniformly with replacement from the current population, and the candidate with the highest validation accuracy is selected.
The first selected parent is excluded before the second tournament, ensuring that the two parents are distinct.
Ties within a tournament are resolved by the first maximal candidate encountered in the sampled tournament ordering.

\begin{figure}[h]
    \centering
    \includegraphics[width=0.95\linewidth]{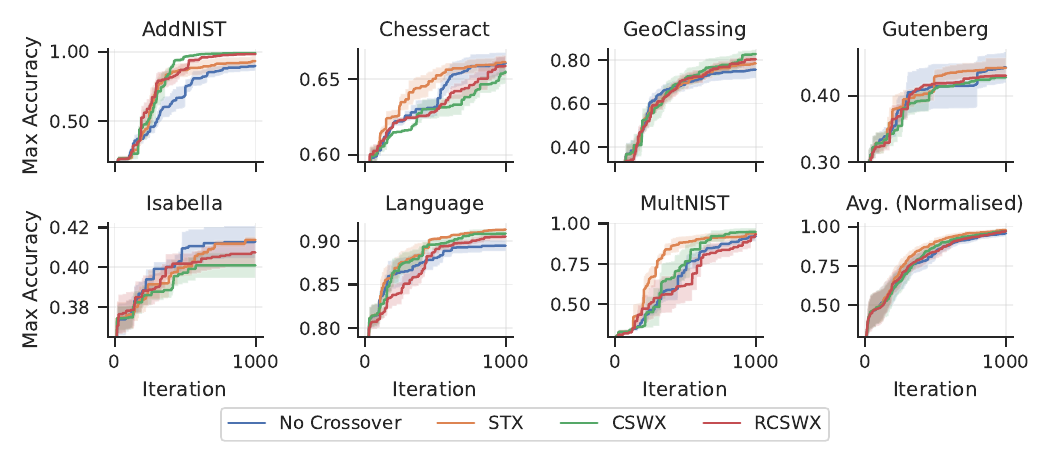}
    \caption{Search results comparing STX, CSWX and RCSWX-based evolutionary search with mutation-only searches. The plots show the mean validation accuracy across \review{five} seeds and the standard error of the mean. To assess the average performance, we normalise the results based on the lowest and highest attained performance within each dataset and combine them into a single plot by averaging (bottom right).}
    \label{fig:search_results}
\end{figure}

Figure~\ref{fig:search_results} demonstrates the attained validation performances throughout the search and Table~\ref{tab:test_scores} shows the final test scores.
They highlight the importance of \review{choosing an appropriate search strategy to explore each individual space: some datasets make good use of the information sharing enabled by crossover---e.g., using CSWX on AddNIST or STX on Chesseract---while others benefit from a less constrained exploration---e.g., mutation-only searches on Isabella.
In some cases, relying on RCSWX's permutation invariant interpolation underperformed compared to the ones using CSWX, which inject some noise into the shortest path in the form of unnecessary mutation operations.
All methods achieved similar validation behaviour on average.} 
The mutation-only condition has the lowest mean test score, although the across-dataset comparison does not establish a statistically significant difference.
Across the seven datasets, the Friedman test did not reject the null hypothesis of equal method ranks ($p = 0.0997$), and none of the Wilcoxon signed-rank comparisons remained significant after Holm correction (Appendix~\ref{app:significance_testing}).
We therefore do not conclude that any method outperforms the others.
Instead, the results show that CSWX and RCSWX remain competitive as search operators while additionally providing edit paths and architecture distances.

\begin{table}[h]
    \centering
    \caption{Test performances of the best models found on the validation set during search, reporting mean and standard deviation across \review{five} seeds. Significance testing can be found in Appendix \ref{app:significance_testing}.}
    \label{tab:test_scores}
    \resizebox{\textwidth}{!}{%
    \begin{tabular}{llllllll|l}
        \toprule
        Dataset       & AddNIST       & Chesseract   & GeoClassing  & Gutenberg    & Isabella     & Language     & MultNIST     & Avg. \\
        \midrule
        No Crossover & 82.13 ± 11.46  & 60.10 ± 0.78  & 79.05 ± 2.44  & 43.35 ± 1.49  & 48.79 ± 2.25  & 93.59 ± 1.06  & 87.99 ± 3.29  & 70.71 ± 1.79 \\
STX           & 97.07 ± 0.30  & 60.91 ± 0.49  & 86.27 ± 2.03  & 42.77 ± 1.07  & 49.94 ± 3.81  & 96.16 ± 0.61  & 91.68 ± 2.41  & 74.97 ± 0.73 \\
CSWX          & 90.64 ± 4.12  & 58.80 ± 0.82  & 82.80 ± 3.29  & 45.75 ± 1.37  & 53.25 ± 1.83  & 95.95 ± 0.41  & 88.94 ± 2.89  & 73.73 ± 0.93 \\
RCSWX         & 95.82 ± 0.36  & 59.94 ± 0.52  & 85.90 ± 2.58  & 44.84 ± 0.41  & 47.27 ± 3.48  & 95.41 ± 0.55  & 91.87 ± 1.22  & 74.44 ± 0.66 \\
        \bottomrule
    \end{tabular}%
    }
\end{table}

\vspace{-20pt}

\vspace{-10pt}




\subsection{Scalability of Constrained Smith-Waterman Crossover}\label{scalability}

Section~\ref{exp1} evaluates whether CSWX and RCSWX remain competitive as search operators at the architecture sizes encountered in the full evolutionary runs.
Here, we address a different question:
whether RCSWX can recover the edit paths of the closest permutation-aware baseline, SEPX, at substantially lower computational cost.
We sort architecture pairs sampled from the searches in Section~\ref{exp1} according to their number of nodes and measure the runtime of the original SEPX, CSWX, and RCSWX implementations.
CSWX is included as the order-sensitive variant, allowing us to distinguish the base sequence-alignment cost from the additional recursive permutation handling in RCSWX.
We additionally report timings for a serial Rust reimplementation of RCSWX; implementation details are given in Appendix~\ref{appendix:big_models}.

\begin{figure}[tbh]
  \centering
  \includegraphics[width=0.9\linewidth, trim=0 11 0 0, clip]{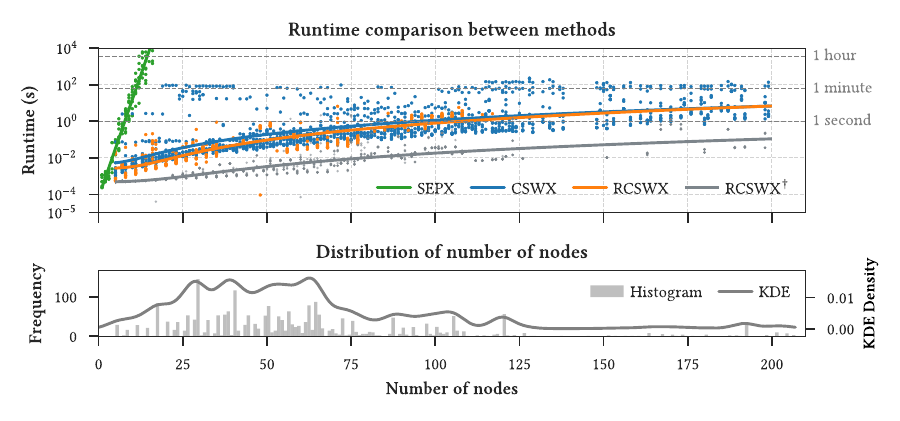}
\vspace{-5pt}
    \caption{
      Runtime versus node count (top; logarithmic runtime axis) and node-count distribution encountered during search (bottom).
      Points show recorded timings, including exception timings and resource-limit lower bounds for Rust.
      Curves are empirical least-squares fits in log-runtime: exponential for SEPX and offset power-law $t_0+a n^b$ for CSWX and both RCSWX implementations.
      Curves extrapolate beyond observed sizes to 200 nodes.
      RCSWX$^\dagger$ denotes the serial Rust reimplementation, measured in aggregate process CPU seconds.
      The campaigns are not matched-hardware or common-clock comparisons.
    }
  \label{fig:cswx-sepx-times}
\end{figure}

For every benchmark pair for which SEPX returned an edit path, RCSWX returned the same path under the matched addition, deletion and substitution costs.
This is an empirical agreement result for the tractable benchmark pairs;
it is not a direct comparison of complete NAS search trajectories.
A full SEPX search comparison is infeasible at the architecture sizes encountered in Section~\ref{exp1} because SEPX becomes computationally intractable before reaching the typical model sizes in those searches.
STX and mutation-only search are not included in this runtime comparison because they do not compute the permutation-aware edit path or architecture distance targeted by this experiment.
We therefore do not claim that CSWX or RCSWX is computationally cheaper than these simpler operators.


Figure~\ref{fig:cswx-sepx-times} shows that SEPX runtime increases sharply and becomes impractical at approximately 20 nodes in this benchmark.
In contrast, the original CSWX and RCSWX implementations complete in less than one second for many of the evaluated architecture pairs;
the first observed computation exceeding one second for the original RCSWX implementation occurs at 71 nodes.
For the Rust implementation, RCSWX$^\dagger$, nearly all successful distance calls in the displayed range use less than one second of process CPU time.
Thus, RCSWX extends permutation-aware edit-path computation into the range of architecture sizes commonly encountered in the searches of Section~\ref{exp1}.
Although its worst-case runtime remains exponential in the number of simultaneously nested branches, the measured results indicate that this factor is not the dominant computational bottleneck for the architecture pairs evaluated here.

These results establish two points.
First, RCSWX empirically reproduces the SEPX edit paths obtained on the tractable benchmark subset under the matched edit costs.
Second, it computes such paths at architecture sizes for which SEPX is no longer practical.
They do not establish that RCSWX improves predictive performance over STX or mutation-only search;
that separate question is evaluated in Section~\ref{exp1}, where no method significantly outperforms the others.

\vspace{-3pt}
\subsection{Smoothness of the Search Space and Population Diversity}\label{smoothness}
As shown above, RCSWX acts as an efficiently computable metric on the search space of architectures. To highlight the benefits of this we use it to analyse the structure, and in particular smoothness, of the architectural loss landscape. We calculate the distance between every pair of the 1000 architectures generated for a single seeded search run on the CIFAR10 dataset\review{, which is a relatively simple and well studied dataset, and on the Isabella dataset, that contains fewer, more complex data samples}. The $1000 \times 1000$ matrices of distances, along with the precomputed performances of said models, have been used to generate the following plots.
For a population $P_t=\{a_1,\ldots,a_N\}$ at search iteration $t$, we define population diversity as the arithmetic mean of the RCSWX distance over all unordered pairs of architectures.

\begin{figure*}[h]
    \centering
    \begin{subfigure}{0.32\linewidth}
        \centering
        \includegraphics[width=\linewidth]{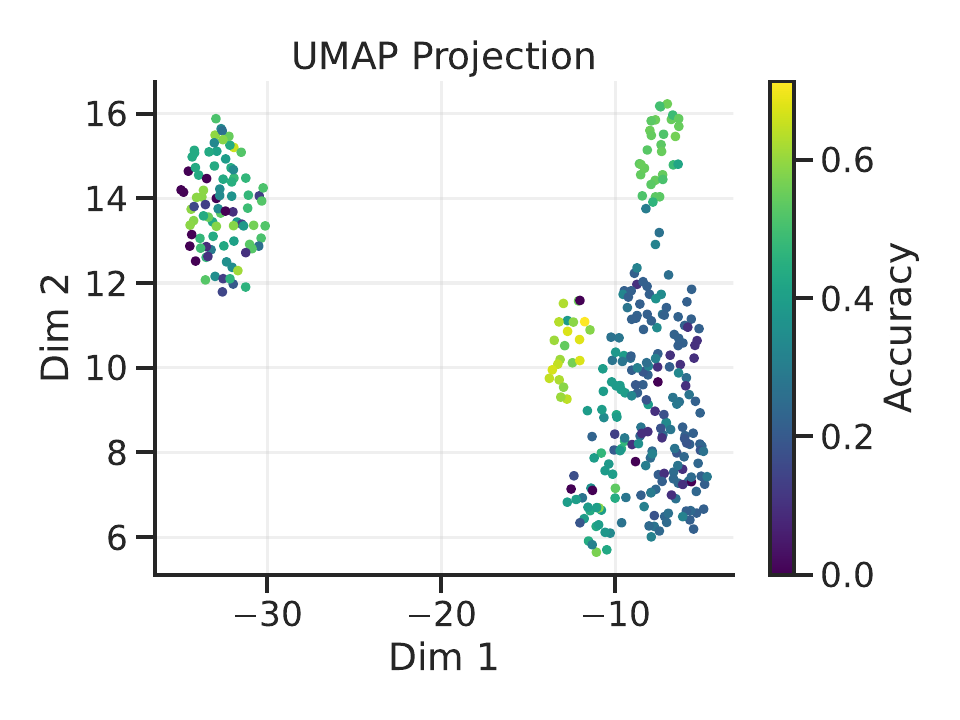}
        \caption{UMAP projection of architectures generated for the CIFAR10 dataset showing clustered structure.}
        \label{fig:umap}
    \end{subfigure}
    \hfill
    \begin{subfigure}{0.32\linewidth}
        \centering
        \includegraphics[width=\linewidth]{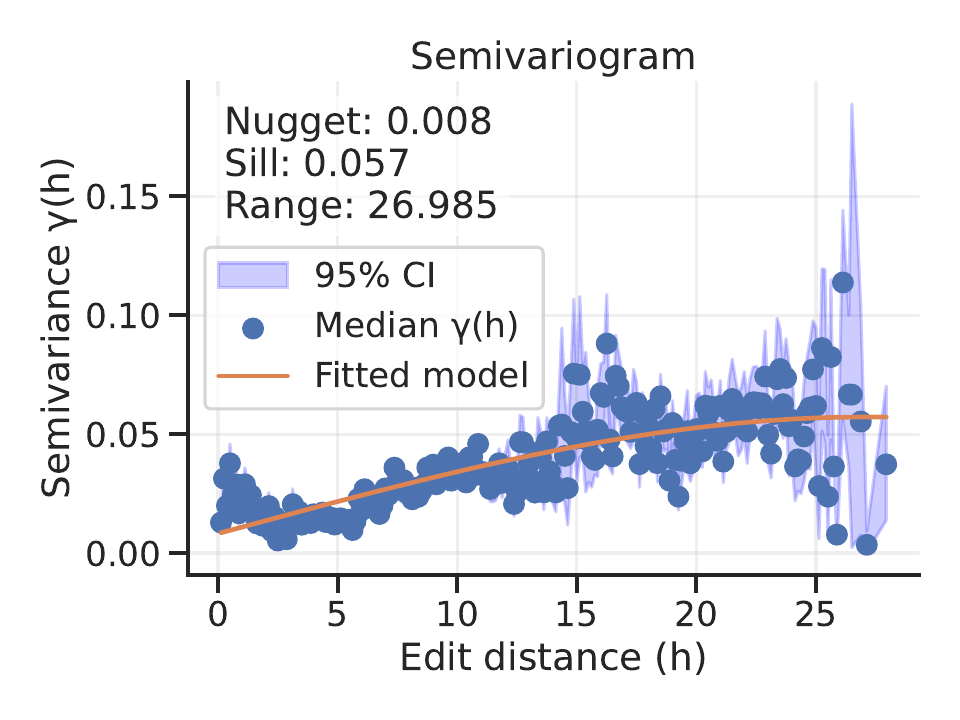}
        \caption{Semivariogram of the population generated for the CIFAR10 dataset  with spherical fit.}
        \label{fig:semivariogram}
    \end{subfigure}
    \hfill
    \begin{subfigure}{0.32\linewidth}
        \centering
        \includegraphics[width=\linewidth]{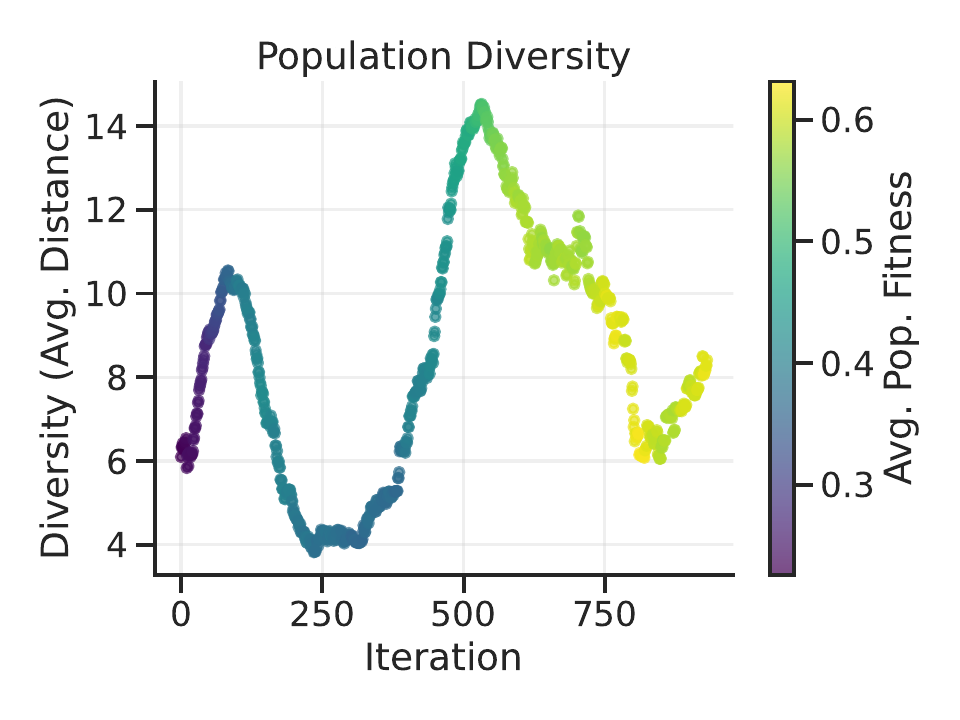}
        \caption{The diversity of the population generated for the CIFAR10 dataset during the search.}
        \label{fig:pop_div}
    \end{subfigure}
    
    \begin{subfigure}{0.32\linewidth}
        \centering
        \includegraphics[width=\linewidth]{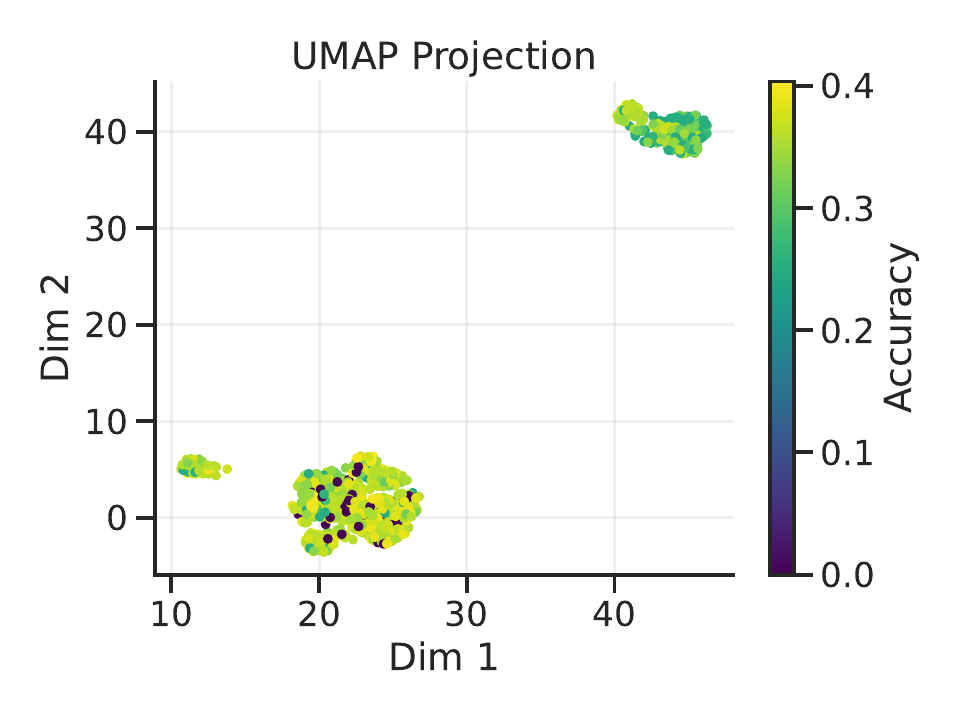}
        \caption{UMAP projection of architectures generated for the Isabella dataset showing clustered structure.}
        \label{fig:umap2}
    \end{subfigure}
    \hfill
    \begin{subfigure}{0.32\linewidth}
        \centering
        \includegraphics[width=\linewidth]{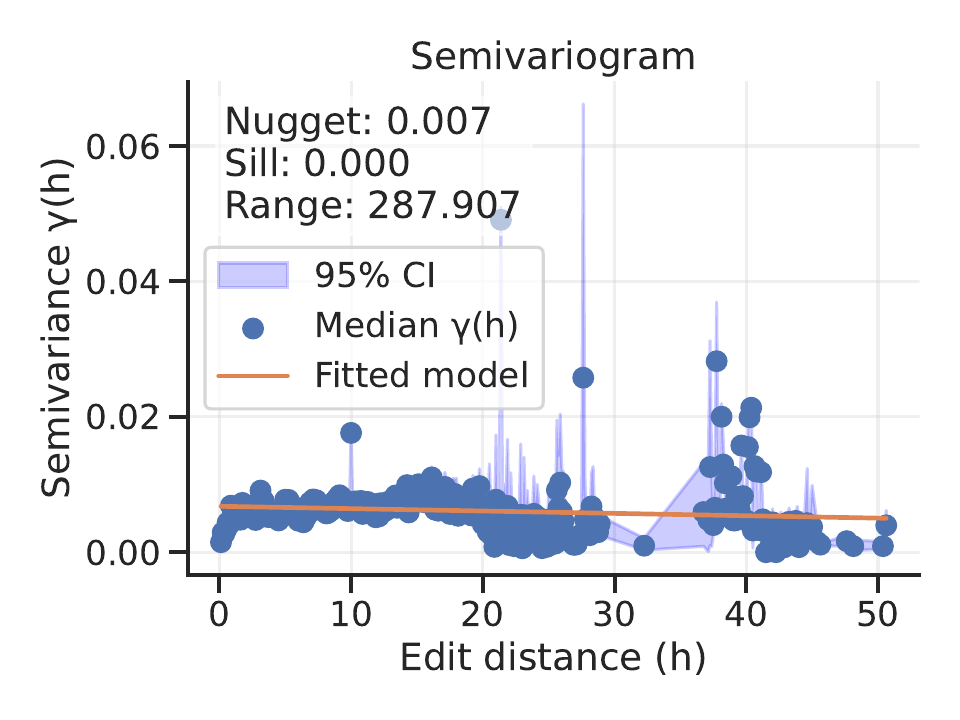}
        \caption{Semivariogram of the population generated for the Isabella dataset with spherical fit. \vspace{1.2em}}
        \label{fig:semivariogram2}
    \end{subfigure}
    \hfill
    \begin{subfigure}{0.32\linewidth}
        \centering
        \includegraphics[width=\linewidth]{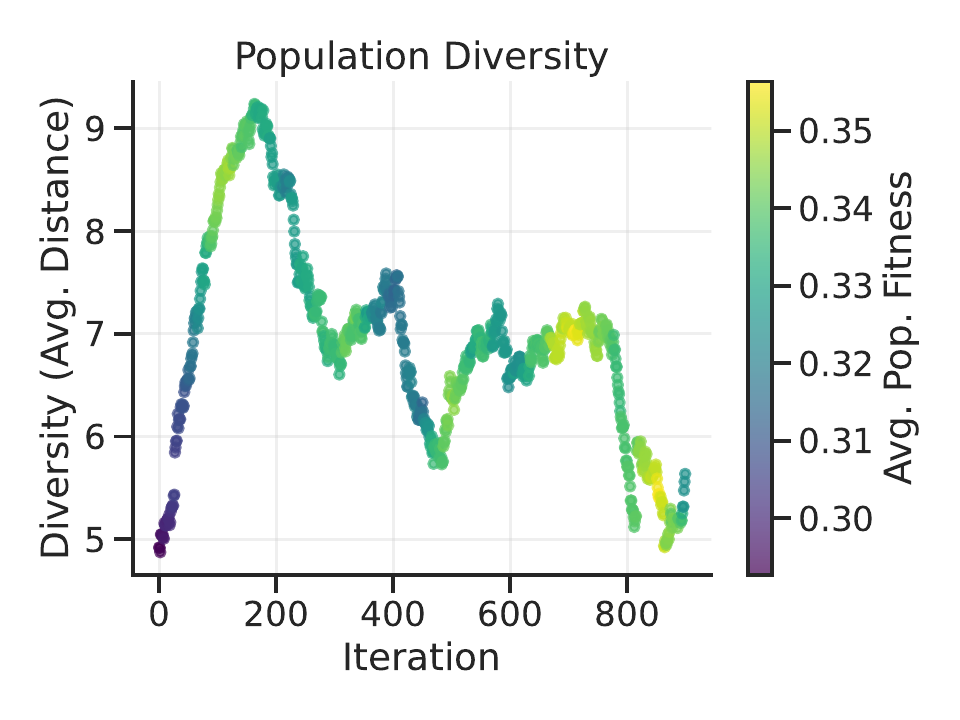}
        \caption{The diversity of the population generated for the Isabella dataset during the search. \vspace{1.2em}}
        \label{fig:pop_div2}
    \end{subfigure}
    \vspace{-4pt}
    \caption{
    Architecture-landscape and population-diversity analyses for CIFAR10 (a--c) and Isabella (d--f).
    Panels (a,d) show two-dimensional UMAP projections based on the RCSWX edit-distance matrices.
    Panels (b,e) show empirical semivariograms and their fitted spherical models.
    Panels (c,f) show population diversity over the course of search, measured as the arithmetic mean of the RCSWX distance over all unordered architecture pairs;
    point colour denotes mean population fitness.
    }
    \label{fig:smoothness_analysis}
\end{figure*}

Figures~\ref{fig:umap} and ~\ref{fig:umap2} show two-dimensional UMAP projections based on the edit-distance matrix. The projections reveal that the explored search spaces are highly fragmented: architectures cluster into a small number of well-separated islands. Within each cluster, models are close in edit distance and tend to exhibit somewhat similar performance, but there is less evidence of continuity across clusters. This suggests that while local neighbourhoods may be smooth, the global search space is disconnected. \review{Some noise can be observed within clusters in both UMAPs, which may correlate to destructive mutations---e.g., disruption of functional blocks, excessive feature condensation, etc---that yield low performing models even within promising regions of the architecture spaces.}


To quantify smoothness more formally, we compute semivariograms in Figures~\ref{fig:semivariogram} and ~\ref{fig:semivariogram2}, which measure how performance similarity decays with increasing edit distance. A semivariogram characterises this relationship through three parameters: the \emph{nugget}, capturing variation at zero distance (often due to noise); the \emph{sill}, corresponding to the maximal variance reached; and the \emph{range}, the distance beyond which no correlation remains. In \review{the case of architectures generated to deal with the CIFAR10 dataset}, for small edit distances ($h \leq 5$), the semivariance $\gamma(h)$ is very low, indicating that close neighbours differ little in performance. As the distance increases, semivariance rises steadily before starting to plateau around $h \approx 15$. Fitting a spherical model yields a small nugget (consistent with low evaluation noise), a sill matching the maximal performance variance, and a range of roughly 25 edits. This implies that performance correlation is maintained within a neighbourhood of this size, but disappears beyond it. \review{Regarding the Isabella dataset, the correlation appears to be maintained throughout the generated population, indicating that very distinct architectures are expected to attain similar accuracies.}


Finally, we use the RCSWX distance metric to track changes in population diversity as each search progresses.
Figures~\ref{fig:pop_div} and~\ref{fig:pop_div2} show visible fluctuations in mean pairwise diversity over the course of the searches.
Increases and decreases in this quantity are consistent with alternating periods of population expansion and concentration;
we interpret these patterns descriptively as exploration- and exploitation-like behaviour rather than as statistically significant phase changes.
The ability to measure and potentially control this dynamic diversity may support future search algorithms in large grammar-based spaces.

\vspace{-4pt}
\section{Discussion and Conclusion}

We have introduced an efficient algorithm for computing distances and performing crossover in grammar-based NAS search spaces, based on a constrained variant of the Smith-Waterman algorithm.
More generally, the method applies to grammar-based spaces that admit a reversible token representation and validity-preserving edit constraints.
In our NAS experiments, it achieves substantial computational gains relative to SEPX while retaining competitive search performance.
The search experiments support competitiveness rather than consistent superiority.
Our distinctive contribution is a combination of a scalable crossover mechanism with a distance measure that enables landscape and diversity analysis.

When used as a distance metric, RCSWX enables landscape analysis at unprecedented scale. Figure~\ref{fig:smoothness_analysis} reveals that both CIFAR10 and Isabella search spaces are locally smooth, in the sense that small mutations tend to produce architectures of similar quality, but differ markedly in global structure. CIFAR10 is highly fragmented, with clusters separated by large distances and little cross-cluster performance correlation, while Isabella is much flatter. This has direct implications for search: crossover-based and hill-climbing methods exploit local neighbourhoods effectively, but escaping to distant high-performing clusters may require restarts or more exploratory strategies.

CSWX enables crossover-based optimisers in grammar-based spaces, while RCSWX adds permutation invariance and a principled distance metric. All methods achieve comparable search performance across datasets, with no single strategy dominating consistently, consistent with prior work arguing that small architectural changes rarely produce large performance shifts~\citep{NASishard,approxsearch}. Notably, CSWX occasionally outperforms RCSWX: imperfect alignment introduces structured noise that broadens exploration, mimicking the behaviour of STX. Our focus here is to introduce a theoretically motivated, computationally efficient tool for the community, rather than to optimise any particular search strategy.
RCSWX reduces repeated computation by retaining permutation-conditioned alignment states, and therefore trades runtime for additional peak memory in deeply nested architectures.

Future work can use the distance metric to explicitly control the 
exploration--exploitation trade-off, and the shortest edit paths to assign component-level merit within architectures, opening a new dimension for discovery and reuse of performant architectural blocks in grammar-based NAS.
Beyond NAS, the same alignment framework may be applicable to other grammar-generated structures for which the serialisation is reversible and domain-valid edits can be enforced. This includes typed program and robot behaviour trees, as well as grammar-based molecular representations with suitable canonicalisation and chemical-validity constraints, discussed in detail in Appendix~\ref{appendix:broad_app}.


\section*{Acknowledgements}
This work was supported in part by Instituto de Salud Carlos III through the projects PMPTA22/00121 and PMPTA22/00118, co-funded by the European Union ‘NextGenerationEU’/PRTR, by Ministerio de Ciencia e Innovación, and the Pro CNIC Foundation, by the Innovationspool of the BMBF-Project: Data-X—Data reduction for photon and neutron science, and by the Engineering and Physical Sciences Research Council [EP/X020703/1].

\bibliography{local}

\newpage
\appendix
\section{Example intermediate offspring}\label{appendix:intermediate_offspring}
In this section, we show the two original models and four offspring architectures---as represented by their derivation trees---that could be generated from the CSWX example in Figure~\ref{fig:matrix}.

\begin{figure}[H]
    \centering
    \subfloat[Parent 1]{\includegraphics[width=0.45\linewidth]{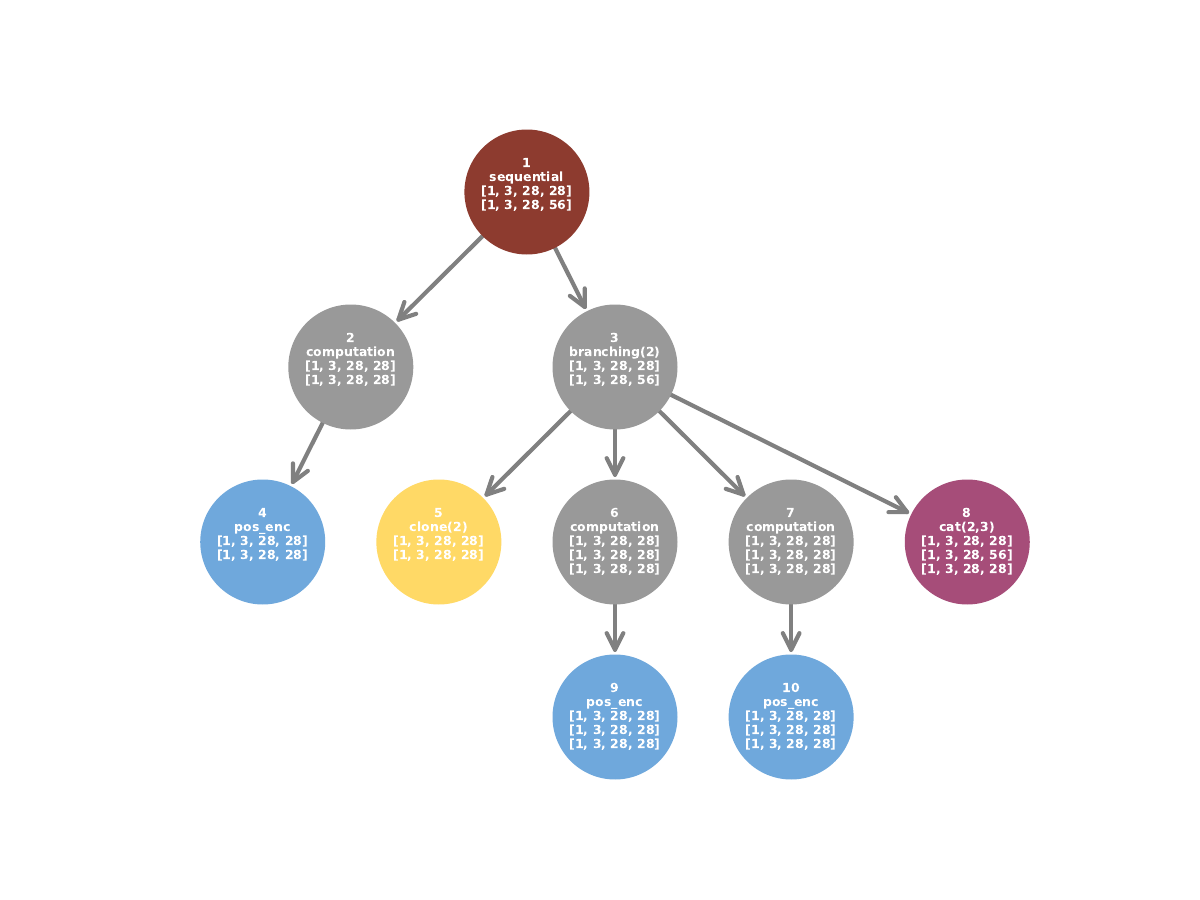}}
    \subfloat[Offspring 1]{\includegraphics[width=0.45\linewidth]{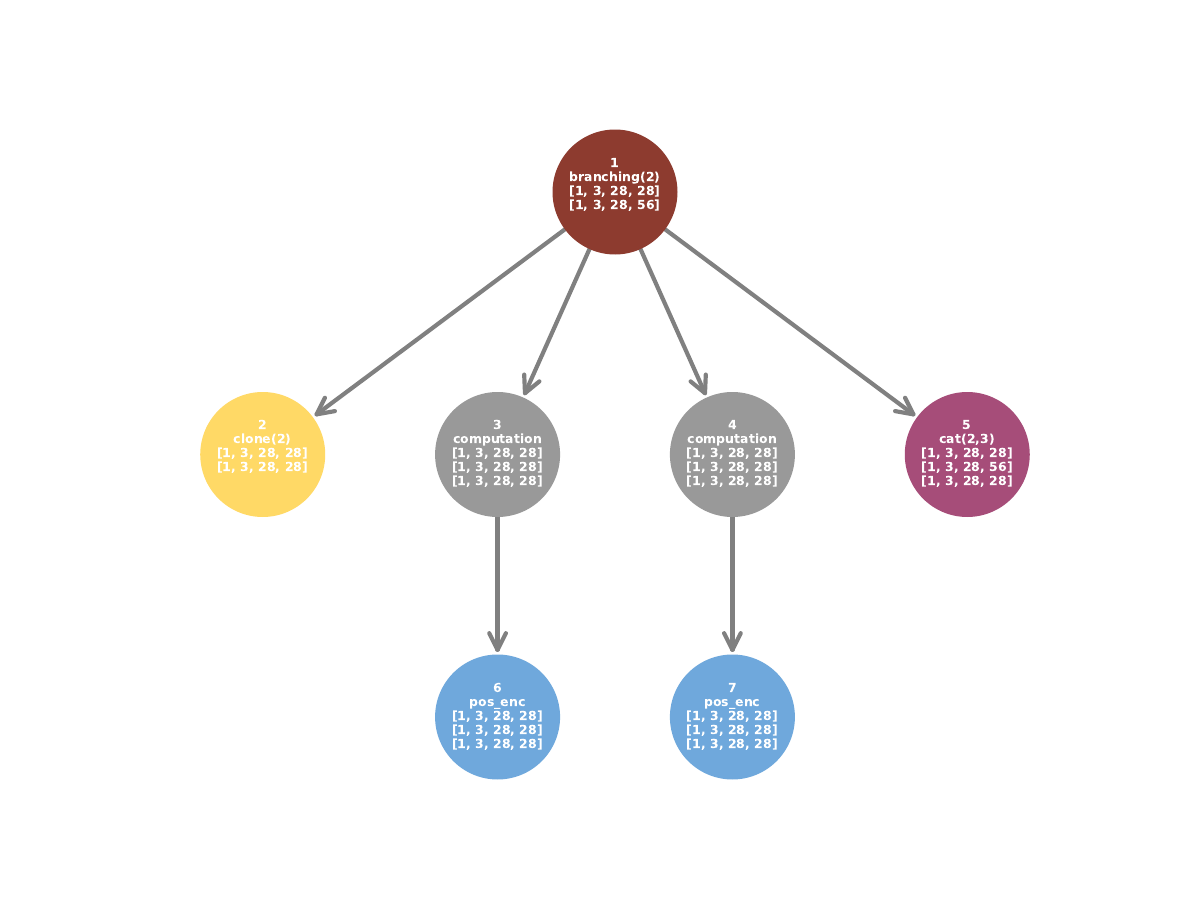}}
    
    \subfloat[Offspring 2]{\includegraphics[width=0.45\linewidth]{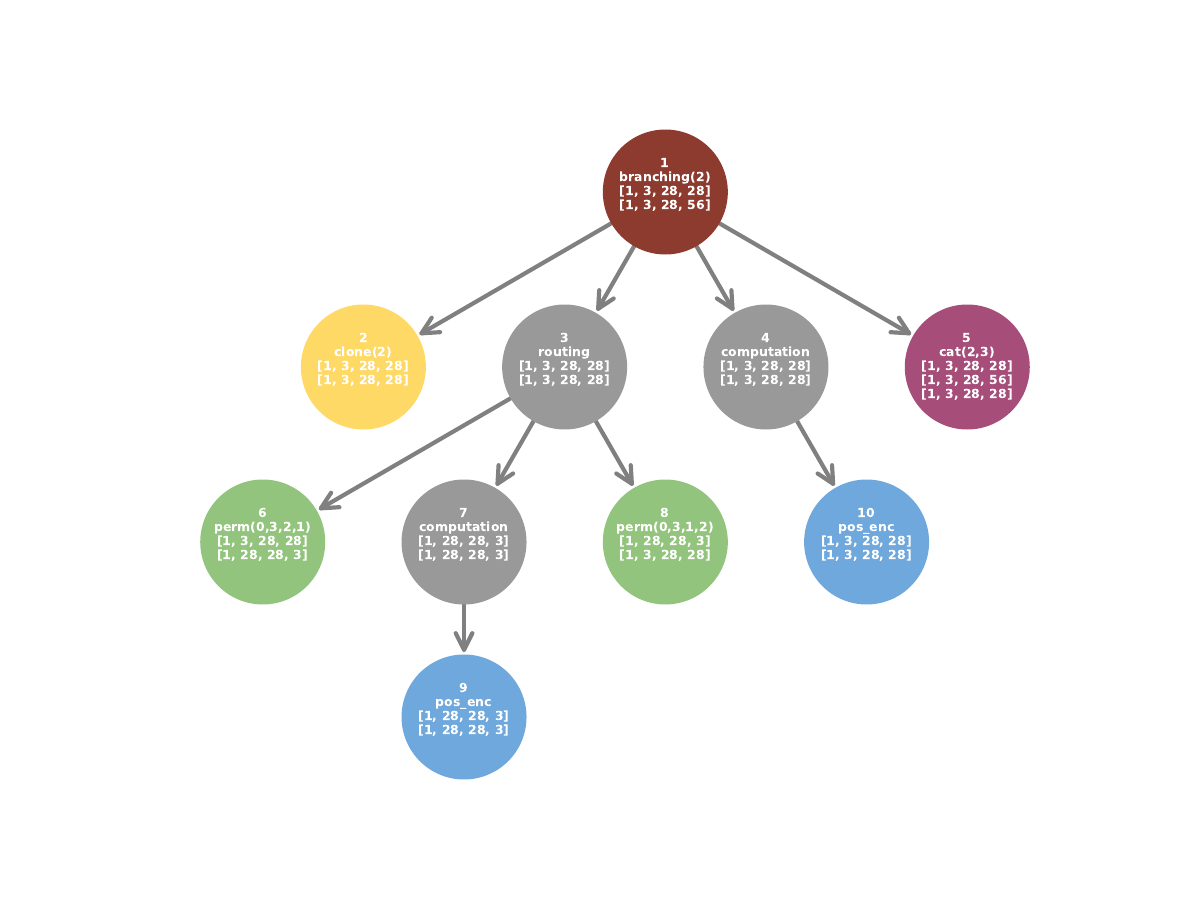}}
    \subfloat[Offspring 3]{\includegraphics[width=0.45\linewidth]{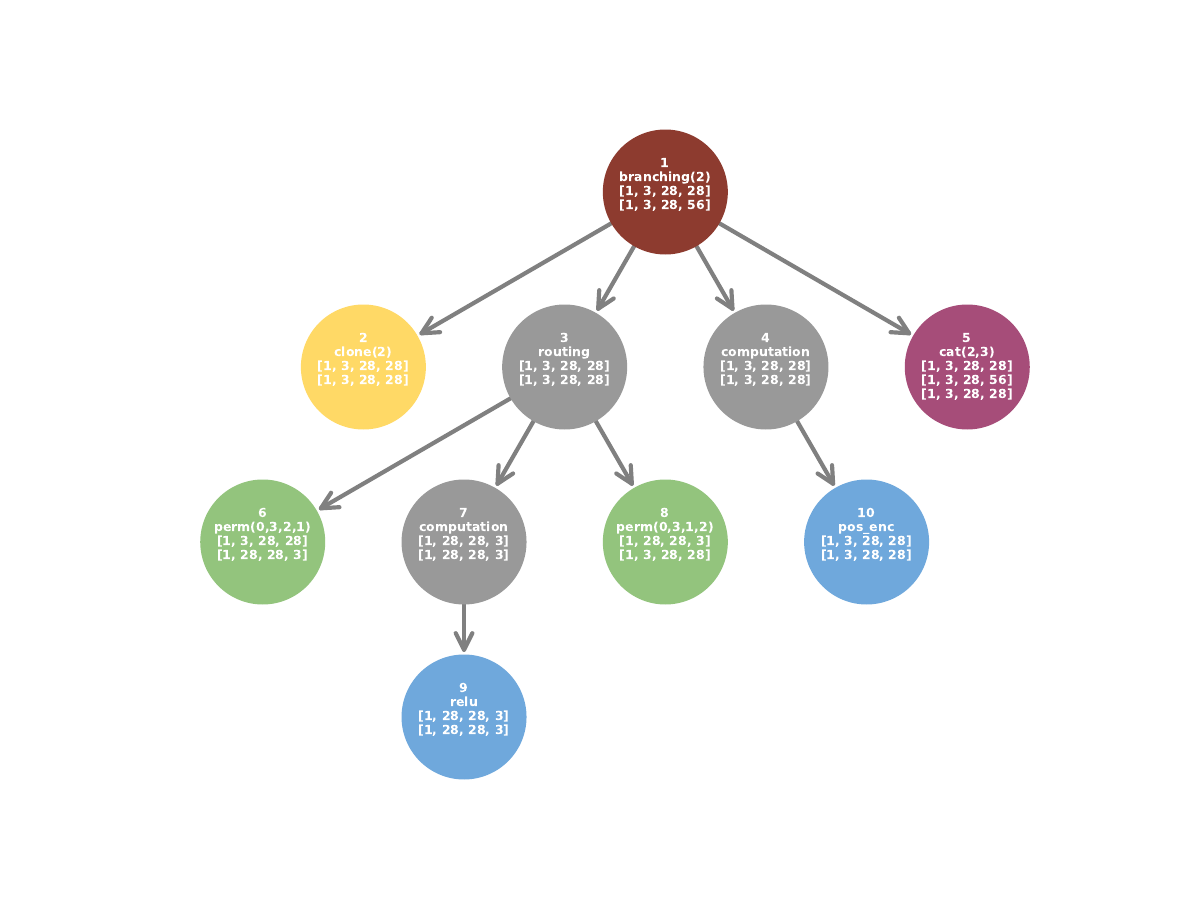}}
    
    \subfloat[Offspring 4]{\includegraphics[width=0.45\linewidth]{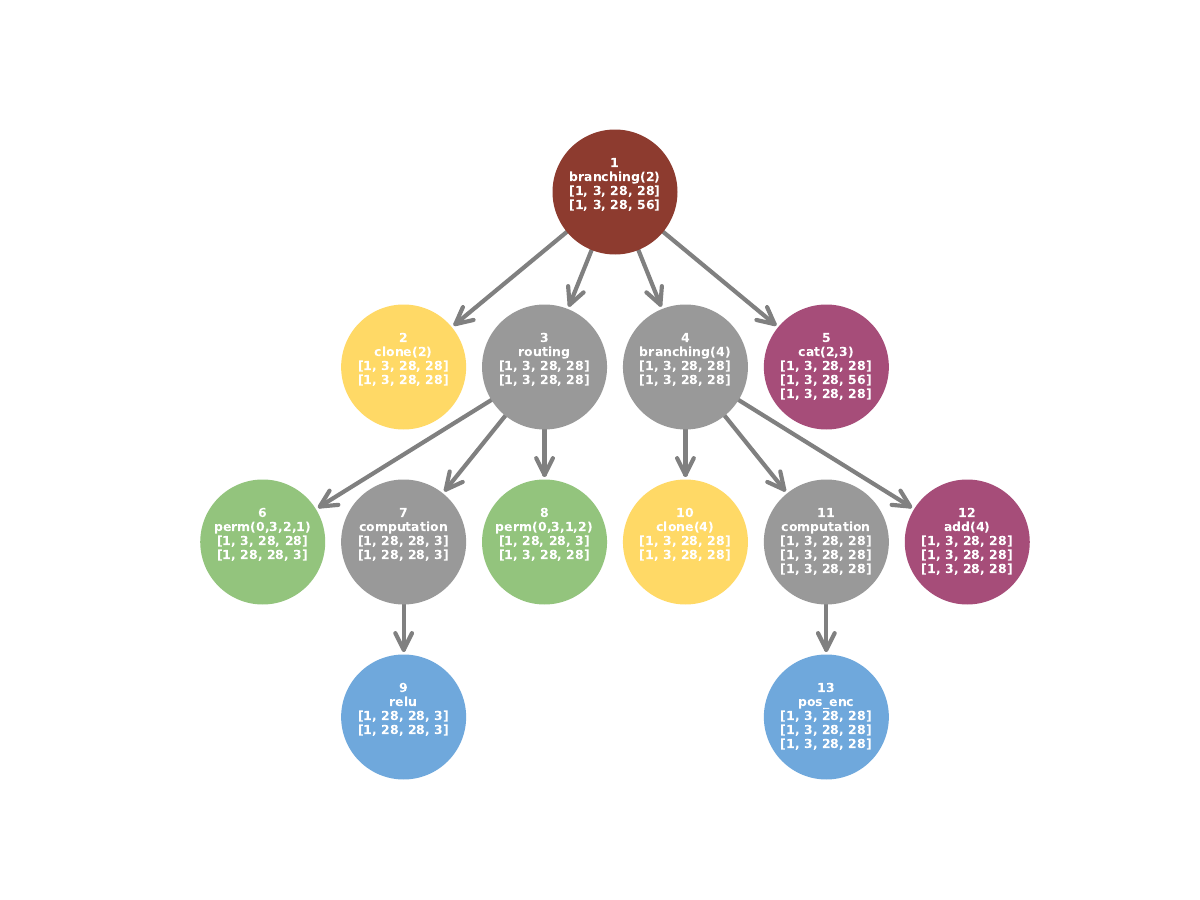}}
    \subfloat[Parent 2]{\includegraphics[width=0.45\linewidth]{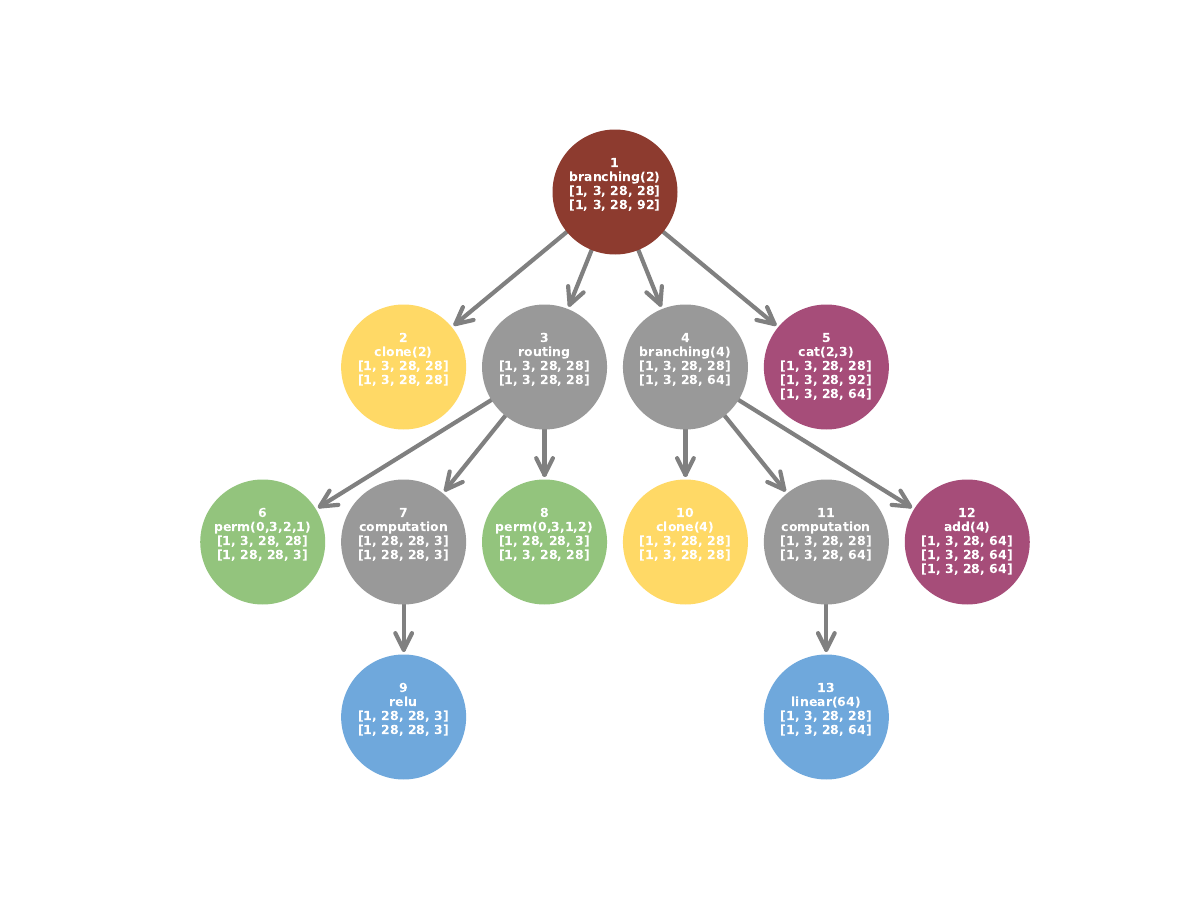}}
    \caption{From left to right, top to down: original model 1 (a), and offspring models resulting from the deletion of the leftmost Computation node (b), the addition of a Routing node (c), the substitution of a positional encoding into a ReLU terminal node (d), the addition of a Branching(4) node (e) and substitution of a positional encoding into a linear terminal node (f), resulting in original model 2.}
    \label{fig:crossover_architectures_offspring}
\end{figure}

\section{Alternative crossover methods}\label{appendix:alt_methods}
In this section, the two methods used to compare against throughout the paper are explained in detail. Both methods successfully generate hybrid offspring, although they differ significantly in complexity and scope.

\subsection{Subtree Crossover}

Subtree crossover (STX) is a classical genetic programming operator that exchanges entire subtrees between parent derivation trees.
\cite{koza1994genetic} demonstrated that this approach preserves coherent functional units, which are essential for maintaining effective computational building blocks.
Transferring whole subtrees encourages the reuse of well-performing substructures and supports modular, hierarchical problem solving.
Furthermore, its ability to handle variable-length representations allows for a more scalable exploration of complex search spaces.
In grammar-based NAS, it is crucial that the exchange subtrees produce syntactically valid architectures.
This is achieved by restricting the crossover operation to subtrees rooted at identical non-terminal symbols.

\vspace{-3pt}
\begin{figure}[ht]
\hspace{-0.05\linewidth}
\begin{minipage}[t]{0.45\linewidth}

\centering
  \captionsetup{width=0.83\linewidth}
  \strut\vspace*{-\baselineskip}\newline\scalebox{0.8}{\input{figures/subtree-crossover-schema}}\vspace{-16pt}
  \caption{Illustration of STX. Parent architectures $\bm{P_1}$ and $\bm{P_2}$ swap selected subtrees (highlighted), generating two offspring architectures $\bm{C_1}$ and $\bm{C_2}$.}
  \label{fig:subtree-crossover}

\end{minipage}%
\hspace{11.5pt}
\begin{minipage}[t]{0.55\linewidth}
\vspace{-4pt}
Figure~\ref{fig:subtree-crossover} on the left provides a visual illustration of STX.
Two parent architectures, $\bm{P_1}$ and $\bm{P_2}$, exchange selected subtrees to yield offspring $\bm{C_1}$ and $\bm{C_2}$.
This procedure swaps portions of the derivation trees while maintaining grammatical validity.
Algorithm~\ref{alg:subtree-crossover} formalises the process.
It begins by extracting node types from each parent and identifying common non-terminal symbols as potential crossover points.
If no common non-terminal exists, the crossover is not performed.
Otherwise, a common non-terminal type is selected at random, followed by one random occurrence of that type in each parent.
The subtree rooted at the selected occurrence in the second parent then replaces the selected subtree in the first parent.
This approach preserves key structural components and ensures that the resulting architecture adheres to the prescribed grammar.

\end{minipage}
\end{figure}

\begin{algorithm}[H]
\caption{Subtree Crossover}
\label{alg:subtree-crossover}
\KwIn{\hspace{10pt}$model1$, the derivation tree for the first architecture to cross over\\
\hspace{40pt}$model2$, the derivation tree for the second architecture to cross over}
\KwOut{\hspace{2pt}$\offspring$, the derivation tree for the resulting hybrid architecture}
\vspace{10pt}
$model1_{types} \gets$ the type of each node in ($model1$)\\
$model2_{types} \gets$ the type of each node in ($model2$)\\
$common_{types} \gets
\{\,t \mid t \in model1_{types} \cap model2_{types},\;
t \text{ is a non-terminal symbol}\,\}$\\
\If{$common_{types} = \emptyset$}{
  \Return \textbf{failure} (no common non-terminals)\\
}
Randomly select a non-terminal type $t \in common_{types}$\\
$pos_1 \gets$ the position of a random occurrence of $t$ in $model1_{types}$\\
$pos_2 \gets$ the position of a random occurrence of $t$ in $model2_{types}$\\
$node_1 \gets$ \textbf{ExtractSubtree}($model1$, $pos_1$)\\
$node_2 \gets$ \textbf{ExtractSubtree}($model2$, $pos_2$)\\
$\offspring \gets$ \textbf{ReplaceSubtree}($model1$, $pos_1$, $node_2$)\\
\Return $\offspring$\\
\end{algorithm}

\subsection{Shortest Edit Path Crossover}

Shortest edit path crossover (SEPX)~\citep{qiu2023shortest} is designed to overcome the permutation problem in evolutionary NAS.
Both parent architectures, $model1$ and $model2$, are represented as graphs, where a graph edit operation modifies the graph by inserting, deleting, or substituting a node or an edge.
The graph edit distance (GED) is defined as the minimum total cost (with unit cost per operation) required to transform one graph into an isomorphic copy of the other, so that GED equals the length of the shortest edit path between the two graphs.

Formally, let
$$
ops^* = \argmin_{ops_i \in ops} \sum_{j=1}^{|ops_i|} cost(ops_{i,j}),
$$
be the shortest edit path from $model1$ to $model2$, which contains $d^*$ unique edit operations.
SEPX generates an offspring graph $\offspring$ by applying half of these edits to $model1$, sampled at random out of $ops^*$.
That is,  
$$
\offspring = ops^*_{\pi_r(\lceil d^*/2 \rceil)} \circ ops^*_{\pi_r(\lceil d^*/2 \rceil - 1)} \circ \cdots \circ ops^*_{\pi_r(1)}(model1),
$$

Here, a single random permutation $\pi_r$ of the $d^*$ edit-operation indices is sampled once for each crossover event.
The first $\lceil d^*/2\rceil$ operations in this ordering are selected and applied sequentially in the permuted order.
The permutation is not resampled between operation applications within the same crossover event, and the composition $\circ$ denotes their sequential application.


SEPX first computes the GED between the two parent graphs, obtaining a shortest sequence of edit operations that transforms one parent into an isomorphic copy of the other.
The operation indices are then shuffled once, and the first $\lceil d^*/2\rceil$ operations in the resulting ordering are applied sequentially to the first parent.
By aligning the parent graphs before recombination, the method addresses the permutation problem---where different genotypes encode the same phenotype---and preserves shared structural components.

\begin{algorithm}[H]
\caption{SEPX}\label{alg:SEP_Crossover}
\KwIn{\hspace{10pt}$model1$, the graph representation of the first architecture to cross over\\
\hspace{40pt}$model2$, the graph representation of the second architecture to cross over}
\KwOut{\hspace{2pt}$\offspring$, the graph representation of the resulting hybrid architecture}
\vspace{10pt}

$ops^* \gets$ \textbf{ComputeShortestEditPath}$(model1,model2)$\\
$d^* \gets |ops^*|$\\
$\pi_r \gets$ one random permutation of $\{1,\ldots,d^*\}$\\
$m \gets \lceil d^*/2\rceil$\\
$ops_{selected} \gets
\left(
ops^*_{\pi_r(1)},
\ldots,
ops^*_{\pi_r(m)}
\right)$\\
$\offspring \gets$ \textbf{ApplyOperations}$(model1,ops_{selected})$\\
\Return{$\offspring$}
\end{algorithm}

In summary, SEPX first uses \textbf{ComputeShortestEditPath} to determine a minimum-cost edit sequence.
It then samples one random ordering of the edit indices, retains the first half of that ordering, and sequentially applies the selected operations to the first parent.
Standard crossover is often disrupted by the permutation problem because different orderings of identical substructures can lead to destructive recombination.
By aligning parent graphs using GED, SEPX preserves coherent functional units, making it both a theoretically principled and practically effective method for combining neural architectures.


\subsection{Comparison to other distance/(dis)similarity and crossover operators}

Table~\ref{tab:comparison} compares (R)CSWX with other distance and crossover methods that have been applied in the context of NAS.


\begin{table}[H]
    \centering
    \caption{
    Comparison with other methods for computing architecture distances and performing crossover.
    A dash indicates that permutation-aware crossover is not applicable because the method does not define a crossover operator.
    }
    \label{tab:comparison}
    \resizebox{\linewidth}{!}{%
    \begin{tabular}{l|ccclll}
        \toprule
        Method
        & Distance Metric
        & Crossover Operator
        & \shortstack{Permutation-aware\\crossover}
        & Encoding
        & Spaces
        & Scaling \\
        \midrule
        WL kernel (Weisfeiler--Lehman)
        & \xmark
        & \xmark
        & --
        & Graph
        & Any
        & $O(h(n_1+m_1+n_2+m_2))$ \\

        NWNAS (Needleman--Wunsch)
        & \cmark
        & \xmark
        & --
        & Sequence
        & Chain-based
        & $O(n_1n_2)$ \\

        GED/SEPX (Graph Edit Distance)
        & \cmark
        & \cmark
        & \cmark
        & Graph
        & Cell-based / general DAGs
        & $O(n_1^{n_2})$ \\

        \midrule

        CSWX
        & \cmark
        & \cmark
        & \xmark
        & Ordered Tree
        & Grammar-based
        & $O(n_1n_2)$ \\

        RCSWX
        & \cmark
        & \cmark
        & \cmark
        & Semi-ordered Tree
        & Grammar-based
        & $O\!\left(
        \sum_{i=1}^{n_1}
        \sum_{j=1}^{n_2}
        2^{d_{ij}}
        \right)$ \\
        \bottomrule
    \end{tabular}%
    }
\end{table}



\section{Constrained Smith-Waterman Crossover functions}
\label{appendix:algs}
In this section, a more detailed description of the functions referenced in Algorithm \ref{alg:CSWC} is laid out.

The \textbf{Serialise} function ensures that the resulting list of nodes required to perform CSWX contains the minimal information required while ensuring that the rules of the grammar are preserved. Commencing with a $node_{start}$, it will recursively add the nodes to a list, ignoring Sequential and terminal nodes that can be inferred from the context, and add a $node_{sep}$ after every branch and routing module to signal where it ends. These separator nodes will contain information about the node they are closing and about whether it signals the separation between two branches or the end of the last branch. 
There is almost a one-to-one conversion between the tree and the list representation of the models, save for the order in which the Sequential nodes are nested (which makes no difference in the actual architecture of the represented models).

\begin{algorithm}
\caption{Serialise}\label{alg:Serialise}

\KwIn{\hspace{10pt}$node$, the model or node we want to serialise}
\KwOut{\hspace{2pt}$node_{seq}$, a serialised representation of $node$}
\vspace{10pt}
\uIf{$node$ is the root node}{$node_{seq}  \gets node_{start}$}\lElse{$node_{seq}  \gets \emptyset$}
\If{$node$ is not a terminal node}{
    \If{$node$ is not a sequential node}{
        $node_{seq} \gets node_{seq} \cup \{node\}$
    }
    \For{$child \in$ children of $node$}{
        $node_{seq} \gets node_{seq} \cup$ \textbf{Serialise}$(child)$\\
        \If{$child$ is not a terminal node \textbf{and} $node$ is a branching or routing node}{$node_{seq} \gets node_{seq} \cup node_{sep}$}
        }
    }
\end{algorithm}

The \textbf{SubstitutionCost} function is really straightforward, comparing two nodes to assign a cost of substituting one into the other. 
If the nodes to compare are Computation nodes, the first and last child will be the same---the type of Computation operation---, while for Branching and Routing nodes the first and last children will define the kind of operations performed. For instance, the $node1$ Branching(4){group(1,4), M, cat(1,4)} would have:

\begin{itemize}
    \item \textbf{SubstitutionCost}($node1$, Branching(4){group(1,4), M, cat(1,4)}) $=0$
    \item \textbf{SubstitutionCost}($node1$, Branching(8){group(1,8), M, cat(1,8)}) $=0.25$
    \item \textbf{SubstitutionCost}($node1$, Branching(4){clone(4), M, add(4)}) $=0.5$
    \item \textbf{SubstitutionCost}($node1$, Computation(identity) $=\inf$
    
\end{itemize}

\begin{algorithm}
\caption{SubstitutionCost}\label{alg:MC}

\KwIn{\hspace{10pt}$node1$, the first node to compare\\
\hspace{40pt}$node2$, the second node to compare}
\KwOut{\hspace{2pt}$cost$, the dissimilarity between $node1$ and $node2$}
\vspace{10pt}

\uIf{$node1$ and $node2$ are the same type of node}{
    \uIf{The first and last children of $node1$ and $node2$ are the same type of node}{
        \uIf{The first and last children of $node1$ and $node2$ have the same hyperparameters}{
            $cost \gets 0$
        }\lElse{$cost \gets 0.25$}
    }\lElse{$cost \gets 0.5$}
}\lElse{$cost \gets \inf$}
\end{algorithm}


\begin{algorithm}
\caption{ValidPath}\label{alg:VP}

\KwIn{\hspace{10pt}$paths$, which holds the direction used to reach each alignment-matrix position\\
\hspace{40pt}$model1_{seq}$, the serialised representation of $model1$\\
\hspace{40pt}$model2_{seq}$, the serialised representation of $model2$\\
\hspace{40pt}$i$, the starting first index in the matrix\\
\hspace{40pt}$j$, the starting second index in the matrix\\
\hspace{40pt}$direction$, the direction we attempt to move towards from the starting position
}
\KwOut{\hspace{2pt}$valid$, which signals whether the direction would be a valid operation}
\vspace{10pt}
$node1 \gets model1_{seq}[i]$\\
$node2 \gets model2_{seq}[j]$\\
\uIf{$direction=$ "sub"}{
    \uIf{$node1$ and $node2$ are the same type of node}{
        \uIf{$node1$ and $node2$ are separators}{
            $condition1 \gets$ False\\
            $condition2 \gets$ False\\
            $i \gets i - 1,\:j \gets j - 1$\\
            \While{\textbf{not} ($condition1$ \textbf{and} $condition2$)}{
                \uIf{$paths[i, j] =$ “add”}{$i \gets i - 1$}
                \uElseIf{$paths[i, j] =$ “del”}{$j \gets j - 1$}
                \uElseIf{$paths[i, j] =$ “sub”}{$i \gets i - 1,\:j \gets j - 1$}
                \If{\textbf{not} $condition1$}{$condition1 \gets node1$ is a separator that closes $model1_{seq}[i]$}
                \If{\textbf{not} $condition2$}{$condition2 \gets node2$ is a separator that closes $model2_{seq}[j]$}
            }
            \uIf{$paths[i, j] =$ “sub”}{$valid \gets$ True}
            \lElse{$valid \gets$ False}
        }\lElse{$valid \gets$ True}
    }\lElse{$valid \gets$ False}
}
\end{algorithm}

\begin{algorithm}[H]
\setcounter{AlgoLine}{25}
\uElseIf{$direction=$ "add"}{
    \If{$node1$ is a separator}{
        $depth \gets$ 0\;
        $i \gets i - 1$\;
        \While{$node1$ does not close $model1_{seq}[i]$}{
            \uIf{$model2_{seq}[j-i]$ is the start of a branching or a routing node}{$depth_{change}\gets1$}
            \uElseIf{$model2_{seq}[j-i]$ closes a branch or a routing node}{$depth_{change}\gets-1$}
            \Else{$depth_{change}\gets0$}
            
            \uIf{$paths[i, j] =$ “add”}{$i \gets i - 1$}
            \uElseIf{$paths[i, j] =$ “del”}{
                $depth \gets depth + depth_{change},\;
                j \gets j - 1$
            }
            \ElseIf{$paths[i, j] =$ “sub”}{
                $depth \gets depth + depth_{change}
                i \gets i - 1,\;j \gets j - 1$\;
            }
        }
        \lIf{$depth = 0$}{$valid \gets$ True}
        \lElse{$valid \gets$ False}
    }
    \lElse{$valid \gets$ True}
}
\ElseIf{$direction=$ "del"}{
    \If{$node2$ is a separator}{
        $depth \gets$ 0\;
        $j \gets j - 1$\;
        \While{$node2$ does not close $model2_{seq}[j]$}{
            \uIf{$model1_{seq}[i-j]$ is the start of a branching or a routing node}{$depth_{change}\gets1$}
            \uElseIf{$model1_{seq}[i-j]$ closes a branch or a routing node}{$depth_{change}\gets-1$}
            \Else{$depth_{change}\gets0$}
            
            \uIf{$paths[i, j] =$ “add”}{
                $depth \gets depth + depth_{change},\;
                i \gets i - 1$
            }
            \uElseIf{$paths[i, j] =$ “del”}{$j \gets j - 1$}
            \ElseIf{$paths[i, j] =$ “sub”}{
                $depth \gets depth + depth_{change},\;
                i \gets i - 1,\;j \gets j - 1$\;
            }
        }
        \lIf{$depth = 0$}{$valid \gets$ True}
        \lElse{$valid \gets$ False}
    }
    \lElse{$valid \gets$ True}
}
\end{algorithm}
\vspace{10pt}
The default costs encode an ordinal hierarchy of edit severity relative to the unit cost assigned to inserting or deleting a node.
A cost of $0$ denotes identical operations;
$0.25$ denotes a change confined to the hyperparameters of an otherwise matching operation;
and $0.5$ denotes a larger change in the child-operation type while preserving the enclosing grammar-compatible node type.
Such compatible substitutions remain cheaper than deleting one node and inserting another, whereas substitutions between structurally incompatible node types are assigned infinite cost and are therefore disallowed.
The numerical values are heuristic defaults rather than theoretically unique or optimal quantities, and they are held fixed across all datasets.
Alternative cost functions could instead incorporate grammar-sampling probabilities, empirical operation effects, or other domain knowledge.
Appendix~\ref{appendix:scoring_matrix} examines the robustness of the resulting distances under several alternative scoring schemes.
A brief analysis of the sensitivity of the distance calculation to the actual mutation costs has been carried out in Appendix \ref{appendix:scoring_matrix}.

\vspace{10pt}
The \textbf{ValidPath} function is where the "Constrained" in "Constrained Smith-Waterman Crossover" comes from, discarding paths that would result in models that do not follow the grammar rules.
In the case of attempting the substitution of one node into another, we first check that they are interchangeable. Substituting a Branching by a Computation, for instance, would result in an incongruous model, as the Computation node would not be suitable for holding the Branching node's children; thus, we only allow substitution between nodes of the same type, regardless of their hyperparameters. Note that Branching(2) is different from all other Branching nodes, as it holds four children instead of three, and thus are not interchangeable. If the nodes we want to substitute are both instances of the $node_{sep}$ class, we trace the path back to make sure that we substituted their associated opening nodes.

If we try to add or delete a module, we only have to be careful when dealing with a $node_{sep}$ instance. If that were the case, we need to trace back the operations to ensure that (1) the associated opening node was dealt with with the same operation and (2) that we are not adding or deleting any non-closed Branching nor Routing nodes, nor any non-opened separator ones. 

\vspace{10pt}
The \textbf{TraceBack} function is used to transform the $dists$ and $paths$ matrices into an actual set of operations that we can perform to transform $model1_{seq}$ into $model2_{seq}$.

Starting from the bottom right corner of the matrices, we follow the path until we reach the top left corner and save all the operations that have a cost, along with all the information required to perform them: what nodes are involved, their positions and, importantly, their interactions with the rest of the operations.
\begin{itemize}
    \item Adding a branching or routing module around certain nodes is disabled by deleting all nodes inside, and enabled by adding any other node inside.
    \item Adding a branching or routing module and all nodes inside is disabled by itself, and enabled by adding any of the other inside nodes.
    \item Deleting a node inside a branching or routing is disabled by deleting all nodes inside said branching or routing, and enabled by either adding another node inside or deleting the branching or routing itself.
\end{itemize}
Knowing which operations enable and disable each other is crucial to be able to generate valid architectures that follow the grammar rules.

\begin{algorithm}
\caption{TraceBack}\label{alg:TB}
\KwIn{\hspace{10pt}$model1_{seq}$, the serialised representation of $model1$\\
\hspace{40pt}$model2_{seq}$, the serialised representation of $model2$\\
\hspace{40pt}$dists$, which holds the distance from $model1$ to every intermediate model\\
\hspace{40pt}$paths$, which holds the steps to reach each intermediate model from $model1$}
\KwOut{\hspace{2pt}$ops_{valid}$, the list of operations to transform $model1$ into $model2$}
\vspace{10pt}

$ops_{valid} \gets \emptyset$\;
$i \gets$ length of $model1_{seq}$\;
$j \gets$ length of $model2_{seq}$\;

\While{$i > 0$ \textbf{or} $j > 0$}{
    $op_{new} \gets$ \{\\
        \hspace{50pt}$id \gets$ length of $ops_{valid}$, serving as the operation identifier\\
        \hspace{50pt}$type \gets$ the kind of operation ``add\_node", ``parallelise", ``substitute"...)\\
        \hspace{50pt}$value \gets$ the cost of performing the operation\\
        \hspace{50pt}$i \gets$ the first index where the node starts\\
        \hspace{50pt}$j \gets$ the second index where the node starts\\
        \hspace{50pt}$ii \gets$, the first index where the node's separators start, if any\\
        \hspace{50pt}$jj \gets$, the second index where the node's separators start, if any\\
        \hspace{50pt}$ops_{disabler} \gets ops_{disabler}$, the operations that forbid performing $op_{new}$\\
        \hspace{50pt}$ops_{enabler} \gets ops_{enabler}$, the operations that allow performing $op_{new}$\\
        \hspace{40pt}\}\\
    \If{$op_{new}\{value\} > 0$}{$ops_{valid} \gets ops_{valid} \cup op_{new}$}
    \If{a branching or routing module has been added or deleted}{
    \For{all operations dealing with nodes contained within}{
    Update their disabler and enabler operations}}
    
    \uIf{$paths[i, j] =$ ``add''}{$i \gets i - 1$}
    \uElseIf{$paths[i, j] =$ ``del''}{$j \gets j - 1$}
    \ElseIf{$paths[i, j] =$ ``sub''}{
        $i \gets i - 1,\;j \gets j - 1$\;
    }
}
\end{algorithm}

\vspace{10pt}
The \textbf{SelectOperations} function takes in all possible operations we can perform to substitute one parent model into another. It first validates each combination of operations, using their sets of disabler and enabler operations to check that no incongruous model would be created. It then assigns each combination a distance from the first model given by the value of each operation within. We then generate a truncated Gaussian probability distribution\review{---optionally presenting a $skewness$ parameter---}that accommodates only values ranging from 0 to the maximum possible distance given by our operations. The probability of each combination of operations will then be drawn from the defined distribution and used to sample one $ops_{selected}$ at random out of all valid combinations.

By default, the $skewness$ parameter is set to zero, generating a truncated, non-skewed Gaussian probability distribution. It could however be defined based on, for instance, the relative performance of both parents. This $skewness$ parameter will make the $\offspring$ produced by $ops_{selected}$ resemble more closely the desired parent. Having a high skewness towards the best out of the two parents would enhance exploitation and reduce exploration, as the models generated would be closer to said parent. It might be interesting to increase this skewness as the search progresses and more promising regions in the space of architectures are found; however, considering that the exploitative RCSWX-based search strategies often underperformed when compared to exploration-focused approaches---see Figure \ref{fig:search_results}---having a high skewness at the beginning would bias the search too much towards the first decent architectures found, hindering overall search performance. In any case, having a high skewness towards the worst performing parent would not make much sense, as it would induce higher exploitation of the least promising regions in the architecture space, benefiting from neither exploration nor thorough exploitation. The influence of the skewness parameter should be less noticeable the closer the parent models are, as these would belong to the same region of the architecture space and not have novel architectures to explore in between the parents.

\begin{algorithm}
\caption{SelectOperations}\label{alg:SelectOperations}

\KwIn{\hspace{10pt}$ops_{valid}$, the set of possible operations\\
\hspace{40pt}$skewness$, the skewness of the operation sampling probability distribution}
\KwOut{\hspace{2pt}$ops_{selected}$, a chosen subset of $ops_{valid}$}
\vspace{10pt}

$combinations \gets \emptyset$\\
\For{$i \leftarrow 1$ \KwTo ($2^{\text{length of }ops_{valid})}$}{
    $combo_{str} \gets$ binary representation of $i$ with as many digits as operations in $ops_{valid}$\\
    $ops \gets$ a subset of $ops_{valid}$ given by $combo_{str}$\\
    \If{$ops$ is a valid set of operations}{$combinations\{combo_{str}\} \gets \sum ops_j\{value\}$}
}

$distr \gets$ a distribution with given $skewness$, truncated to 0 and $\max(combinations)$\\ 
$probs \gets$ probability of each combination's value given by $distr$\\
$probs \gets probs \slash \sum probs$\\
$combo_{selected} \gets $ an operation subset from $combinations$ sampled with probability $probs$\\
$ops_{selected} \gets$ a subset of $ops_{valid}$ given by $combo_{selected}$
\end{algorithm}

\section{Robustness to Alternative Scoring Schemes}
\label{appendix:scoring_matrix}

The numerical scoring scheme affects the absolute edit distances and may also affect which alignment path is selected.
To assess the robustness of the distance calculation to this modelling choice, we compare the default scoring scheme with three alternatives using 50 pairs of randomly generated architectures.
The alternatives vary both the substitution costs and, in one case, the treatment of branching-node insertion and deletion, as summarised in Table~\ref{tab:scoring_matrices}.
This analysis evaluates broad robustness to alternative scoring schemes rather than independently tuning the default values $0.25$ and $0.5$.

\begin{table}[!htbp]
    \centering
    \caption{
    Default and alternative scoring schemes used to assess the robustness of the RCSWX distance calculation.
    }
    \label{tab:scoring_matrices}
    \begin{tabularx}{\linewidth}{p{35pt}p{115pt}X}
        \toprule
        Name & Addition/deletion cost & Mutation cost\\
        \midrule
        \midrule
        Scoring matrix 0 & Always 1 & 0 if nodes are the same, 0.25 if nodes' children's hyperparameters differ, 0.5 if nodes' children's type of node differ and $\infty$ if parents type of node differ\\
        \midrule
        S. M. 1 & Always 1 & 0 if nodes' children's are exactly the same, 0.5 if not, $\infty$ if parent's type of node differ\\
        \midrule
        S. M. 2 & Always 1 & $\infty$ if parent's type of node differ, 0.5 otherwise\\
        \midrule
        S. M. 3 & Number of branches for branching nodes, 1 otherwise & Difference in number of branches for branching nodes, same as default scoring matrix otherwise\\
        \bottomrule
    \end{tabularx}
\end{table}

\begin{figure}[tbh!]
  \centering
  \includegraphics[width=\linewidth*4/5]{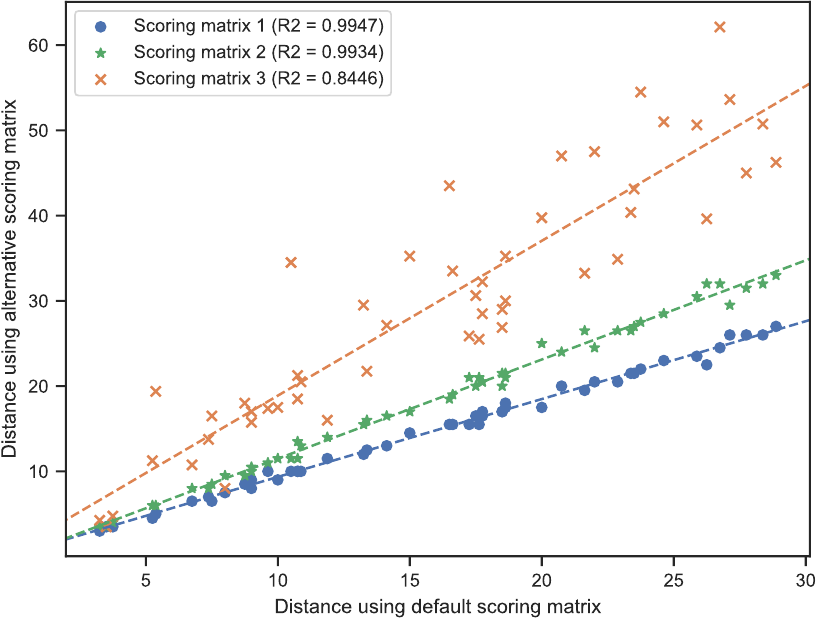}
  \caption{
      Distance calculated for randomly sampled pairs of models employing scoring matrices 0 (x axis) and 1 through 3 (y axis) with RCSWX, along with their linear fits and R$^2$.
  }
  \label{fig:scoring_matrices}
\end{figure}

All models were sampled from \textit{einspace}, having a length ranging from 4 to 104 layers, sampled uniformly. Each architecture was paired with another one with the same length. The results are depicted in the Figure \ref{fig:scoring_matrices}.

Figure~\ref{fig:scoring_matrices} shows that changing the scoring scheme changes the absolute RCSWX distances, as the relationships are not perfectly linear.
The two moderate alternatives retain an almost linear relationship with the default distances ($R^2=0.9947$ and $R^2=0.9934$), whereas the
branch-dependent scoring scheme produces greater deviation ($R^2=0.8446$).
Thus, global distance values are largely stable under moderate changes to the scoring scheme, but become more sensitive when structural edits are reweighted substantially.

This experiment evaluates broad robustness to alternative scoring schemes rather than independently varying the default values $0.25$ and $0.5$.
It therefore does not establish that exact shortest paths, local
nearest-neighbour relations, crossover offspring, or complete search trajectories are invariant to the scoring scheme.
We regard the scoring matrix as a domain-dependent modelling choice:
the default scheme provides a simple normalised hierarchy of edit severity, while applications requiring a different notion of architectural similarity may define alternative costs.

\section{Crossing over predesigned models}
\label{appendix:big_models}
In this section, we explore the capabilities of RCSWX to compare and cross over relatively big predesigned models. We have constructed members of the ResNet and MLP-Mixer families in the \textit{einspace} grammar and compared them, obtaining the following results.

\paragraph{Optimized implementation}
We provide a standalone reimplementation with a safe Rust core and Python
bindings.\footnote{\url{https://github.com/flxai/rcswx}, v0.3, commit
\href{https://github.com/flxai/rcswx/tree/8d4ace02627ad97ed6293b5e6c598ce0ebd05eee}
{\texttt{8d4ace0}}.}
Figure~\ref{fig:cswx-sepx-times} and Table~\ref{tab:big_models}
include runtime measurements for the serial Rust implementation.
The search and landscape experiments use the original implementation.


\begin{table}[tbh!]
    \centering
    \small
    \caption{RCSWX results on predefined architectures. Rust v0.3
    times are medians of three serial alignment wall-time measurements
    with corner collapsing enabled.}
    \label{tab:big_models}
    \begin{tabular}{@{}lrrrr@{}}
        \toprule
        Architectures & Nodes & \shortstack{Original}
        & \shortstack{Rust} & Distance\\
        \midrule
        ResNet18, MLP-Mixer d8  & 267, 264 & 39.054 s   & 0.728 s  & 60.38\\
        ResNet18, MLP-Mixer d12 & 267, 392 & 342.383 s  & 5.529 s  & 95.38\\
        ResNet34, MLP-Mixer d8  & 499, 264 & 86.633 s   & 1.113 s  & 119.62\\
        ResNet34, MLP-Mixer d12 & 499, 392 & 741.57 min & 97.136 s & 115.62\\
        \bottomrule
    \end{tabular}%
\end{table}

The resulting alignment matrices are depicted in the following figures.

\begin{figure}[H]
  \centering
  \includegraphics[width=0.75\linewidth]{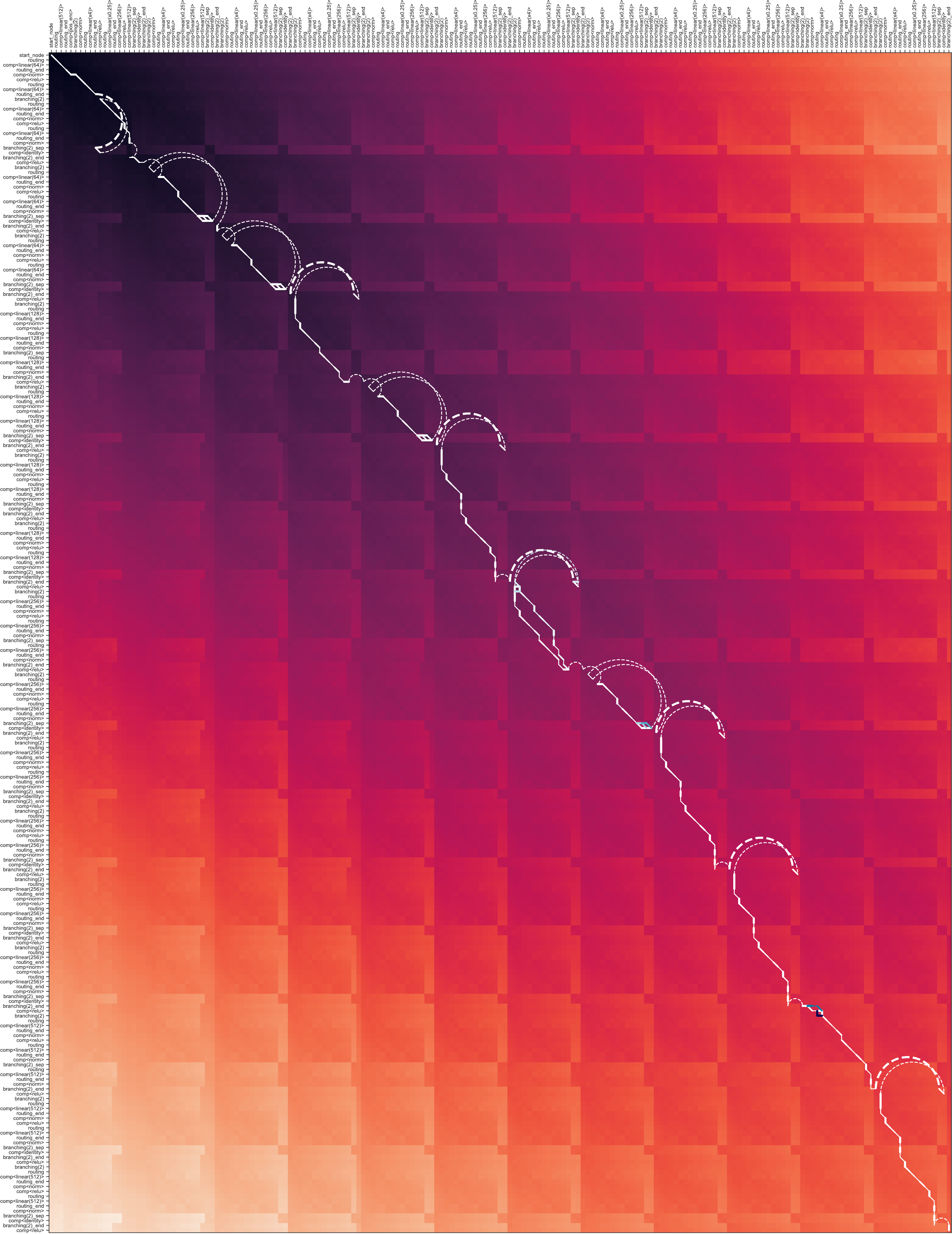}
  \caption{
      Alignment matrix for ResNet34 and MLP-Mixer d12
  }
  \label{fig:resnet34mlp12}
\end{figure}

\begin{figure}[H]
  \centering
  \includegraphics[width=0.75\linewidth]{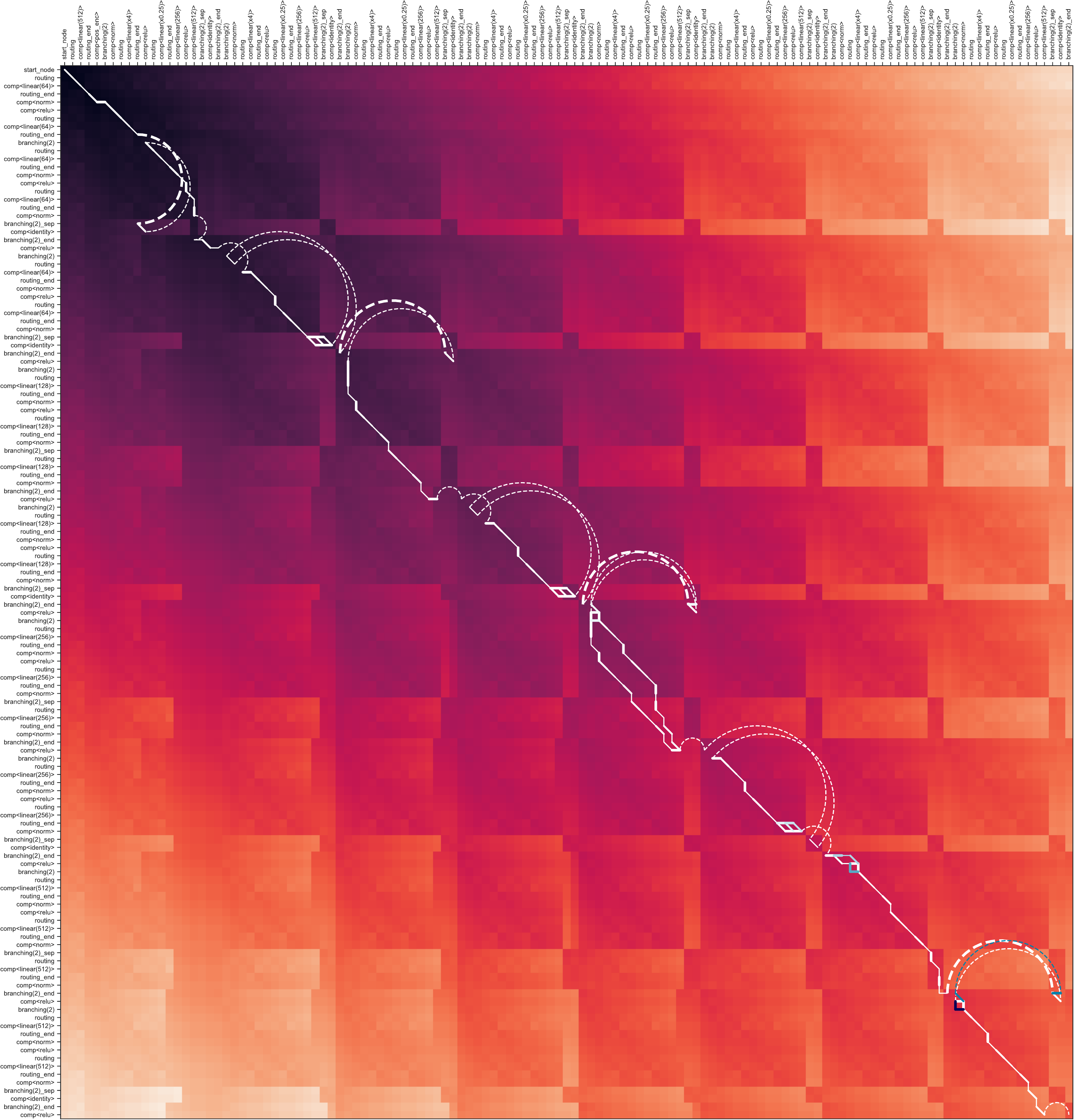}
  \caption{
      Alignment matrix for ResNet18 and MLP-Mixer d8 
  }
  \label{fig:resnet18mlp8}
\end{figure}

\begin{figure}[H]
  \centering
  \includegraphics[width=0.75\linewidth]{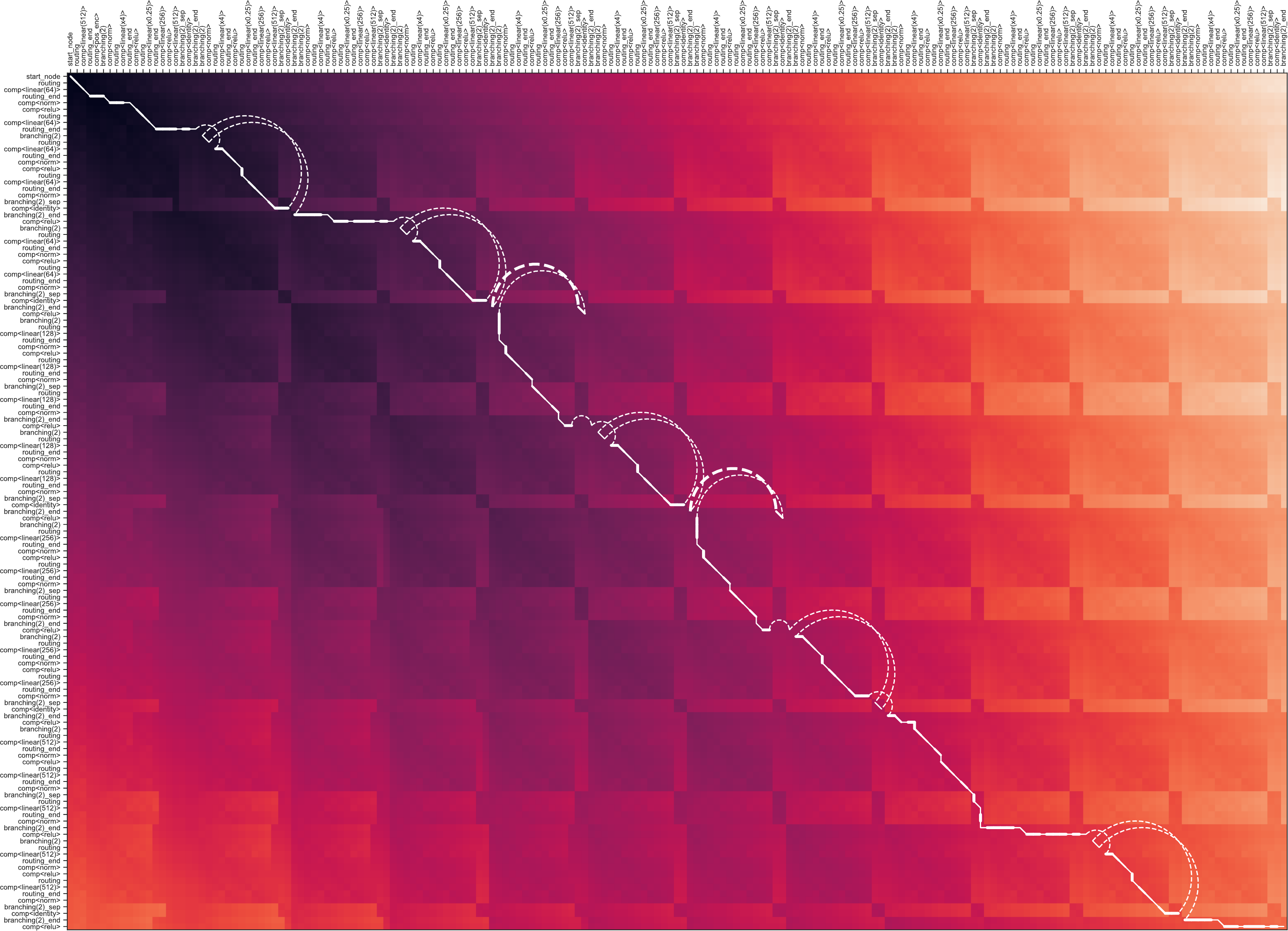}
  \caption{
      Alignment matrix for ResNet18 and MLP-Mixer d12 
  }
  \label{fig:resnet18mlp12}
\end{figure}

\begin{figure}[H]
  \centering
  \includegraphics[width=0.75\linewidth]{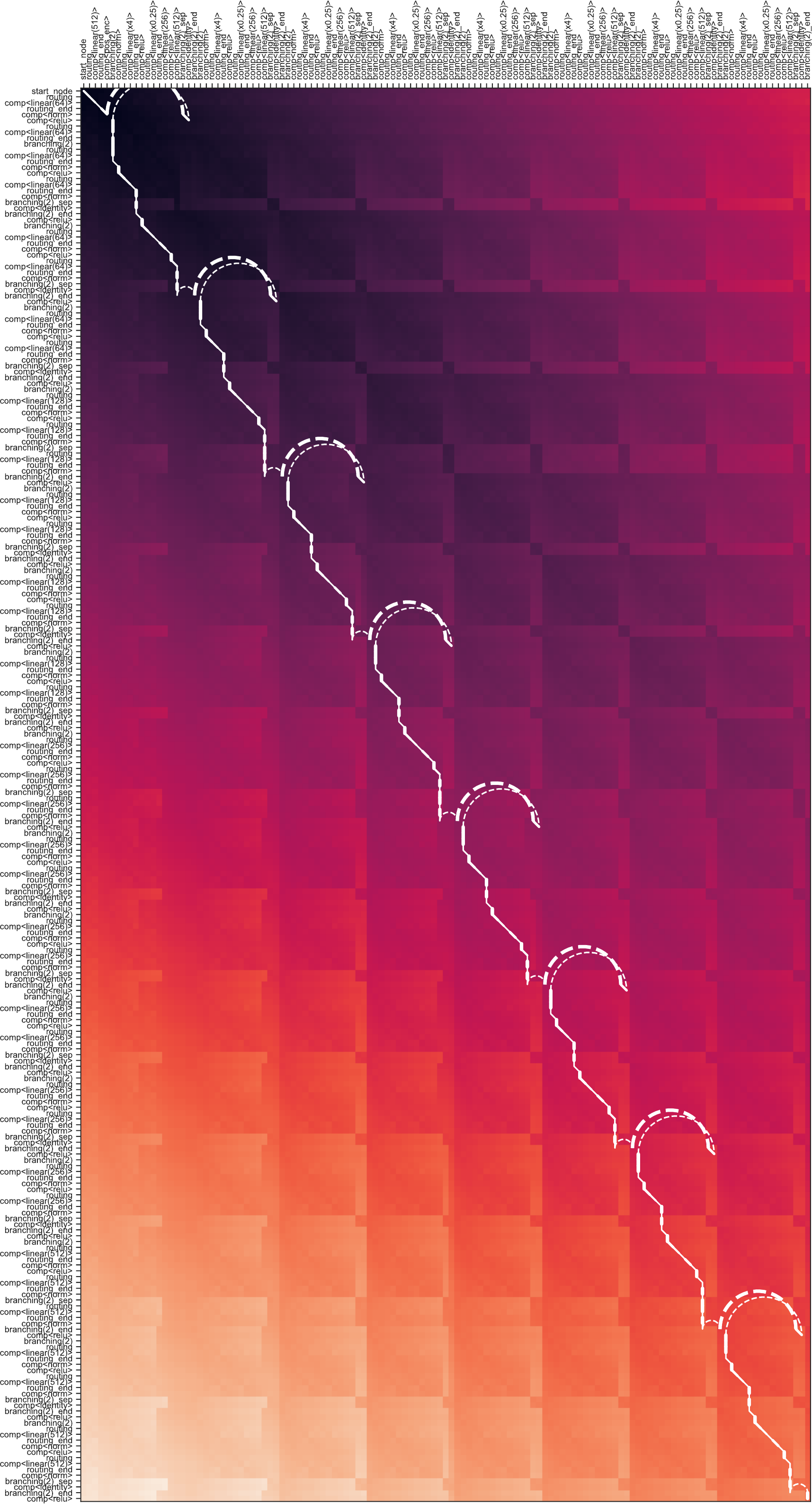}
  \caption{
      Alignment matrix for ResNet34 and MLP-Mixer d8 
  }
  \label{fig:resnet34mlp8}
\end{figure}

The repetitive block-based nature of the architectures becomes apparent when compared, with cyclical patterns---whose actual shapes and cycle lengths are defined by the alignment of the first and final layers of the architectures---spread throughout the shortest edit paths. 
Also note that compute times are relatively low considering the large number of nodes for the aligment of ResNet18 and MLP-Mixer d8. This can be explained by the low depth throughout the matrix. There are no nested branches in either of the models, which lowers the amount of recursive computing---which translates into reduced computations, but also reduced overhead as well. However, compute times are relatively high for the alignment of ResNet34 and MLP-Mixer d12. This is due to redundant, irrelevant paths, mostly in the lower left and upper right corners, as checked empirically after the computations were performed. While alignments surrounding the shortest edit path are relatively constrained, there are a lot of equally low-performing paths further away towards the corners---e.g., it is as costly to mutate the first ResNet block into the first MLP-Mixer block and then add the second MLP-Mixer block than it is to add the first MLP-Mixer block and then mutate the first ResNet block into the second MLP-Mixer block. These paths are redundant in the sense that they achieve the same final distance and obey the same constraints. Keeping track of all these paths is not only memory intensive but compute intensive as well, as every possible path needs to be checked for constraints to continue forwards with the next operation.

The phenomena discussed above highlight the deviation in compute time from theory to practice and suggest that, for real applications with warm initialisations, RCSWX might be way faster than the original timings in Figure \ref{fig:cswx-sepx-times}, given that the redundant and unpromising paths are collapsed or ignored. This collapse can either be cell-wise---each time we compute a cell, we collapse all redundant paths into one---or matrix-wise---considering how unlikely it is to find a best path very far away from the main diagonal of the alignment matrix, we can restrict the maximum number of path calculations at the corners by employing techniques akin to constraining with the Sakoe-Chiba band or the Itakura parallelogram \citep{bands}.

\section{Broader applicability} \label{appendix:broad_app}
In this work we have exemplified the use of (R)CSWX with the implementation of a simple genetic evolution optimiser to explore models expressed in the \textit{einspace} grammar. However, our goal is to provide an efficient tool that enables researchers and NAS practitioners to deploy any crossover-based algorithm on their own search spaces, and even apply (R)CSWX to sequence alignment outside of the Deep Learning field. In this section, we suggest how to use (R)CSWX to construct different optimisers---using two or more parent architectures to generate offspring based both on model performance or relative distances---and work with complex search spaces to demonstrate its wide applicability.

\subsection{(R)CSWX on complex search spaces}
\subsubsection{Multi input-output spaces}
Allowing multiple input and output nodes in an architecture can be achieved by paying with the constraints in the grammar---that is, by modifying the mutation costs.
For spaces where all architectures present the same number of inputs and outputs, which is the most common case, it is enough to give the addition/deletion of input and output nodes---as well as their substitution from or into any other type of node---a cost of $\infty$. This would force (R)CSWX to hinge around these nodes, aligning the intermediate segments.
For cases with varying number of input and output nodes, input and output nodes can be treated as any other type of node as there should not be any actual constraints on their position within the architectures. If we want the best paths to align as many of these nodes as possible, though, it would be sufficient to set the cost of adding/deleting these nodes to a high enough number---for instance, if adding or deleting any other node has a cost of 1, the cost of adding or deleting an input/output node can simply be the length of the biggest parent---and the cost of substituting from or into any other type to $\infty$. An example alignment matrix calculated in this fashion is shown in Figure \ref{fig:multiinpout} below.

\begin{figure}[H]
  \centering
  \includegraphics[width=0.75\linewidth]{figures/multiinpout.pdf}
  \caption{
      Alignment matrix for two sequences with a differing number of input nodes (addition/deletion cost of 1; substitution cost of 0 if nodes are the same, 0.5 if they are interchangeable). Each matrix cell is subdivided into substitution, deletion, addition and selected costs. Selected operations are shown in a light colour, and deprecated operations are shown in grey. The shortest edit path is shown in green.
  }
  \label{fig:multiinpout}
\end{figure}

\vspace{-3pt}

Note that the case with aligned input and output nodes can be treated as a specific, fortuitous, instance of the unaligned multi-input and -output; however, setting the addition/deletion costs to $\infty$ instead of a large value allows to directly flag and ignore all positions in the alignment matrix outside the regions defined by these hinge points, speeding up computations considerably for large models.

\subsubsection{Recursive graphs}
Aligning recursive architectures using the (R)CSWX is not a trivial task. The best way to approach this is to make sure that the sequential representation of the graph is representative, reversible and unique---or, at least, limited in such a way that we can calculate all possible alignments recursively in a reasonable amount of time---and, then, define the mutation costs appropriately to make sure that the constraints of the grammar are respected. Using the input and output nodes as anchor points, we can define the sequence of nodes employing simple graph traversal algorithms, like breadth-first search, depth-first search, or encoder-based approaches, like \citep{xu2018graph2seqgraphsequencelearning, liu2022neighbor2seqdeeplearningmassive, chen2025flattengraphssequencestransformers}. Special separator tokens can be introduced during the definition of the sequences. By carefully defining the addition, deletion and substitution costs for these separator tokens, we can force the shortest edit paths to respect the rules of the space we are exploring---similarly to what has been accomplished with the constraints imposed to the \texttt{Branching(2)} layers in the \textit{einspace} grammar---and/or weight the alignment of the architecture loops according to our needs.

\subsection{(R)CSWX to define search algorithms}
\subsubsection{Particle Swarm Optimisation}
In particle swarm optimisation (PSO)~\citep{kennedy_particle_1995}, each individual in the population calculates its new position in the space by moving in a direction defined by a vector, which is interpolated from the vector pointing towards the best position achieved by the individual so far and the vector pointing towards the best individual in the current population. These two components can be weighted by the fitness of their respective individuals, and the interpolated vector can be rescaled based on a given step size. This approach has already been successfully applied to NAS~\citep{lankford2024neuralarchitecturesearchusing, Turky2026-af}, but its application is heavily hindered by the ability to perform crossovers in the search space defined. We propose a method to perform multi-parent crossover by pooling the mutation operations yielded by (R)CSWX into four sets from which to sample from, allowing the deployment of any multi-parent interpolation-based algorithm, like PSO.

First, the original model is aligned with both parents separately using (R)CSWX.
Then, the two resulting sets of operations are compared with one another.
Those pairs of operations that are exactly the same---e.g., deleting a \texttt{Computation(ReLU)} node from the original individual that is not present in either one of the other two parents---are bagged together in set $A$.
Then, those pairs of operations that are similar---e.g., substituting a \texttt{Computation(ReLU)} node into a \texttt{Computation(linear64)} in one of the alignments and into a \texttt{Computation(linear32)} in another---are bagged together into set $B$.
The rest of the operations are bagged individually in sets $C_1$ and $C_2$, according to the alignment matrix they belong to.

The number---or combine cost---of operations to be performed can be modulated by a step size parameter. We sample operations at random from set $A$ until needed. If we sample all operations from set $A$, we continue sampling pairs of operations from set $B$. The actual operation to perform out of the pair can be sampled according to the individuals' fitness scores. If set $B$ is depleted, we select operations from the remaining sets $C_1$ and $C_2$. Once again, the set from which to sample each operation can be selected probabilistically based on the individuals' respective fitness.

After applying the selected operations, the generated offspring is guaranteed to be closer to both parents simultaneously, given that they share some common nodes. This method can be expanded to perform multi-crossover across any number of parents, but the probability of finding common operations reduces drastically and the definition of further operation sets---e.g., operations shared by all alignments, operations shared by all but one alignments, etc---might be necessary to avoid treating all operations as unpaired.

\subsubsection{Estimation of Distribution Algorithms}
Using the distances across all models in the population, we can define a high dimensional space---similarly to Section \ref{smoothness}---and construct a probabilistic model that shifts the sampling of the new offspring towards the most promising regions of the space. This sampling can be performed by either (1) selecting parents for crossover that define lines that cross over said promising regions and using the skewness parameter to control the sampling while remaining probabilistic or (2) performing multi-parent interpolation similarly to the approach described to perform PSO.

\subsubsection{Differential Evolution}
Differential Evolution (DE) can be implemented by combining the sets of operations yielded by (R)CSWX. A population size $NP\geq 4$ and a crossover probability $CR\in[0,1]$ need to be selected. Due to the space to explore being combinatorial rather than numerical, the definition of a differential weight $F \in [0,2]$ is not trivial and we opt to simply set it to 1.

First, for each individual in the population $x_n$, we select three other individuals $x_1$, $x_2$ and $x_3$, and define the set of operations that transform models $x_a$ and $x_b$ as $\Omega_{x_a, x_b} = x_a - x_b$. We can then calculate a donor set of operations $\Omega_{\text{donor}}$ as $\Omega_{\text{donor}} = \left(x_1 - x_n\right) \cup \left(\left(x_2 - x_n\right) \not\in \left(x_3 - x_n\right)\right)$. A single operation is sampled out of $\Omega_{\text{donor}}$ to mimic the behaviour of the $\delta$ random variable in the DE algorithm. Then, for each other operation in $\Omega_{\text{donor}}$, we discard it with probability $p=1-CR$, generating the set $\Omega_{\text{final}}$. To generate the new individuals, we simply apply all operations in $\Omega_{\text{final}}$ to $x_n$.

\subsubsection{Firefly Algorithm}
In the firefly algorithm, for each search step $t$, each individual $x_i$ in the population of size $NP$ is updated by moving towards each other individual $x_j$ that presents a better fitness based on their distance $r_{ij}$ and a random permutation $\alpha_t \epsilon_t$ such that $x_i^{t+1}=x_i^t+\alpha_t \epsilon_t+\sum_{j=1}^{NP}\beta (x_j^t-x_i^t)e^{-\gamma r_{ij}^2}$. The random permutation can be achieved with a simple mutation, while the term that moves each individual towards the best ones in the population can be calculated using (R)CSWX.

First, for each individual $x_i$ we select every architecture $x_j$ that presents better fitness and calculate the set of operations $\Omega_{j}$ that transforms $x_i$ into $x_j$. Then, for each operation across all models that pertain to a certain node---e.g., one node in $x_i$ is a \texttt{Computation(linear64)} that is either deleted or substituted to a different type of \texttt{Computation} node in some of the alignments---we select the one pertaining to the alignment with individual $x_j$ with a probability proportional to $e^{-\gamma r_{ij}^2}$, generating a set of operations $\Omega_{\text{selected}}$. These distances $r_{ij}$ are already known as it is given by (R)CSWX, which has already been used to generate the sets of mutation operations. Lastly, each operation in $\Omega_{\text{selected}}$ is either selected or discarded based off a probability $\beta$, generating $\Omega_{\text{final}}$. By applying the operations in $\Omega_{\text{final}}$, we produce an offspring architecture that is interpolated from the population based on both their relative fitnesses and distances.

\subsubsection{Custom Diversity-driven Algorithms}
The distance metric provided by RCSWX can be employed to explicitly control the exploration–exploitation tradeoff, be it by regularising architecture scores with a diversity term or selecting a novel individual for training out of a batch of mutated ones. This allows for tweaking already existing optimisers or even proposing new ones. Explicitly controlling the diversity of the population is not only expected to further improve results by helping escape local minima, but is also crucial for deep ensemble learning and other scenarios where non-regularised NAS tends to underperform.

\section{Significance Testing} \label{app:significance_testing}
To assess whether the four algorithms---mutation-only, STX, CSWX and RCSWX-driven evolutionary algorithms---differ significantly in performance across datasets, we applied a non-parametric Friedman test to their accuracy ranks in the datasets' test partitions.
The Friedman test did not reject the null hypothesis of equal method ranks ($p = 0.0997$).
This does not establish statistical equivalence;
rather, the experiment does not provide sufficient evidence to distinguish the methods' across-dataset performance.
For completeness, we also conducted Wilcoxon signed-rank pairwise comparisons with Holm correction, with none of the pairwise comparisons recognised as significant.
The corresponding critical difference (CD) diagram reflects this outcome: all algorithms fall within a single cluster, illustrating that their rank differences do not exceed the critical threshold.
These results together indicate that we cannot conclude that any algorithm outperforms the others across the evaluated datasets.

\begin{figure}[h]
    \centering
    \includegraphics[width=0.75\linewidth]{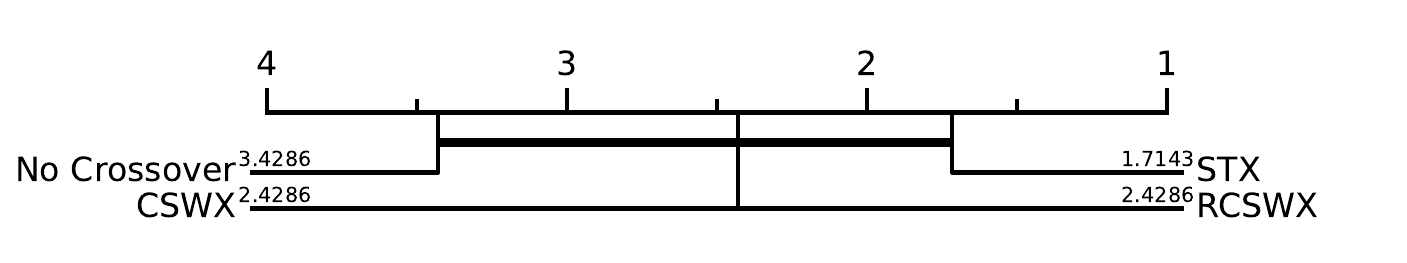}
    \caption{Critical difference (CD) diagram comparing the average ranks of the four algorithms across all datasets. The Friedman test did not reveal a statistically significant difference among the methods (p = 0.0997), and none of the Wilcoxon–Holm post-hoc comparisons reached significance. All algorithms fall within a single cluster—indicated by the horizontal connection—showing that their rank differences do not exceed the critical threshold.}
    \label{fig:cd_diagram}
\end{figure}

\section*{LLM Usage}
Large language models were used only to aid and polish the writing of this paper, as well as auto-complete code fragments.

\end{document}

%% file: math_commands.tex
\usepackage{amsmath,amsfonts,bm}

\def\eqref#1{equation~\ref{#1}}

\def\1{\bm{1}}

\DeclareMathAlphabet{\mathsfit}{\encodingdefault}{\sfdefault}{m}{sl}
\SetMathAlphabet{\mathsfit}{bold}{\encodingdefault}{\sfdefault}{bx}{n}

\DeclareMathOperator*{\argmin}{arg\,min}

%% file: figures/subtree-crossover-schema.tex
\begin{tikzpicture}[>=stealth]

\node (P1node) at (0,0) {
  \begin{forest}
    for tree={
      circle,
      minimum size=8mm,
      edge={-},
      line width=1.5pt,
      draw=black,
      tikz+={remember picture}
    }
    [ {}, label=above:{$\bm{P_1}$}
      [ {}, edge={line width=0.7pt}
        [ {}, draw=none, edge={dotted, black, line width=1.5pt} ]
        [ {}, draw=none, edge={dotted, black, line width=1.5pt} ]
      ]
      [ {}, draw=orange, edge={line width=0.7pt}
        [ {}, draw=none, edge={dotted, orange, line width=1.5pt} ]
        [ {}, draw=none, edge={dotted, orange, line width=1.5pt} ]
      ]
    ]
  \end{forest}
};

\node (P2node) at (4,0) {
  \begin{forest}
    for tree={
      circle,
      minimum size=8mm,
      edge={-},
      line width=1.5pt,
      draw=black,
      tikz+={remember picture}
    }
    [ {}, label=above:{$\bm{P_2}$}
      [ {}, draw=RedViolet, edge={line width=0.7pt}
        [ {}, draw=none, edge={dotted, RedViolet, line width=1.5pt} ]
        [ {}, draw=none, edge={dotted, RedViolet, line width=1.5pt} ]
      ]
      [ {}, edge={line width=0.7pt}
        [ {}, draw=none, edge={dotted, black, line width=1.5pt} ]
        [ {}, draw=none, edge={dotted, black, line width=1.5pt} ]
      ]
    ]
  \end{forest}
};

\node (C1node) at (0,-3) {
  \begin{forest}
    for tree={
      circle,
      minimum size=8mm,
      edge={-},
      line width=1.5pt,
      draw=black,
      tikz+={remember picture}
    }
    [ {}, label=above:{$\bm{C_1}$}
      [ {}, edge={line width=0.7pt}
        [ {}, draw=none, edge={dotted, black, line width=1.5pt} ]
        [ {}, draw=none, edge={dotted, black, line width=1.5pt} ]
      ]
      [ {}, draw=RedViolet, edge={line width=0.7pt}
        [ {}, draw=none, edge={dotted, RedViolet, line width=1.5pt} ]
        [ {}, draw=none, edge={dotted, RedViolet, line width=1.5pt} ]
      ]
    ]
  \end{forest}
};

\node (C2node) at (4,-3) {
  \begin{forest}
    for tree={
      circle,
      minimum size=8mm,
      edge={-},
      line width=1.5pt,
      draw=black,
      tikz+={remember picture}
    }
    [ {}, label=above:{$\bm{C_2}$}
      [ {}, draw=orange, edge={line width=0.7pt}
        [ {}, draw=none, edge={dotted, orange, line width=1.5pt} ]
        [ {}, draw=none, edge={dotted, orange, line width=1.5pt} ]
      ]
      [ {}, edge={line width=0.7pt}
        [ {}, draw=none, edge={dotted, black, line width=1.5pt} ]
        [ {}, draw=none, edge={dotted, black, line width=1.5pt} ]
      ]
    ]
  \end{forest}
};

\draw[->, orange, line width=1.2pt, dashed]
      (1.45, -0.7) -- (2.65, -2.9);
\draw[->, RedViolet, line width=1.2pt, dashed]
      (2.65, -0.7) -- (1.45, -2.9);

\end{tikzpicture}